%% file: main_tech_report.tex
\documentclass[10pt, logo, copyright]{nvidiatechreport}

\usepackage{mdframed}
\usepackage[utf8]{inputenc} % allow utf-8 input
\usepackage[T1]{fontenc}    % use 8-bit T1 fonts

\usepackage{amsfonts}       % blackboard math symbols
\usepackage{nicefrac}       % compact symbols for 1/2, etc.
\usepackage{microtype}      % microtypography
\usepackage[dvipsnames]{xcolor}         % colors
\usepackage{multirow}
\usepackage{multicol}
\usepackage{graphicx}
\usepackage[numbers]{natbib}
\usepackage{tabto}
\usepackage{xspace}
\usepackage{amsmath}
\usepackage{adjustbox}
\usepackage{enumitem}
\usepackage{wrapfig}
\usepackage{dblfloatfix}

\usepackage{footmisc}
\usepackage{listings}
\usepackage{setspace}

\usepackage[table]{xcolor}
\usepackage{float}

\usepackage{hyperref}
\usepackage{array}
\usepackage{ragged2e}
\usepackage{subcaption}
\usepackage{caption}

\usepackage{natbib}
\input{math_commands.tex}

\usepackage{amsmath}
\usepackage{amssymb}
\usepackage{mathtools}
\usepackage{amsthm}
\usepackage{enumitem}
\setlist[itemize]{itemsep=0.5em}
\usepackage{multirow}
\usepackage{booktabs}

\usepackage[capitalize,noabbrev]{cleveref}

\usepackage{algorithm}
\usepackage{algorithmicx}
\usepackage{algpseudocode}
\usepackage{xspace}
\usepackage{csquotes}
\usepackage{listings}
\usepackage{xcolor}
\definecolor{NvidiaGreen}{rgb}{0,0.392,0}

\usepackage{multirow}
\usepackage{multicol}
\usepackage{enumitem}
\usepackage{listings}
\usepackage{xcolor}
\usepackage{tcolorbox}
\tcbuselibrary{listings, skins}
\usepackage{caption}
\DeclareCaptionLabelFormat{nolabel}{}
\definecolor{lightgray}{gray}{0.95} % 背景色
\definecolor{darkblue}{rgb}{0,0,0.6} % 标题颜色
\definecolor{nvgreen}{cmyk}{50, 0, 100, 0}

\definecolor{headerblue}{RGB}{0,0,180}
\definecolor{thinkblue}{RGB}{40,40,200}
\definecolor{toolgreen}{RGB}{0,120,0}
\definecolor{responsepurple}{RGB}{120,0,160}
\definecolor{obsbrown}{RGB}{150,90,20}
\definecolor{boxborder}{RGB}{0,0,160}
\definecolor{boxbg}{gray}{0.97}

\title{GDPO: Group reward-Decoupled Normalization Policy Optimization for Multi-reward RL Optimization}

\author{Shih-Yang Liu\textonesuperior, Xin Dong\textsuperscript{*}, Ximing Lu, Shizhe Diao, Peter Belcak, Mingjie Liu, Min-Hung Chen, Hongxu Yin, Yu-Chiang Frank Wang, Kwang-Ting Cheng\textonesuperior, Yejin Choi, Jan Kautz, Pavlo Molchanov  \\~\\ \textbf{NVIDIA} \\ \ }
\correspondingauthor{X}

\begin{abstract}

As language models become increasingly capable, users expect them to provide not only accurate responses but also behaviors aligned with diverse human preferences across a variety of scenarios. To achieve this, Reinforcement learning (RL) pipelines have begun incorporating multiple rewards, each capturing a distinct preference, to guide models toward these desired behaviors. However, recent work has defaulted to apply Group Relative Policy Optimization (GRPO) under multi-reward setting without examining its suitability. 
In this paper, we demonstrate that directly applying GRPO to normalize distinct rollout reward combinations causes them to collapse into identical advantage values, reducing the resolution of the training signal and resulting in suboptimal convergence and, in some cases, early training failure. We then introduce \textbf{G}roup reward-\textbf{D}ecoupled Normalization \textbf{P}olicy \textbf{O}ptimization (\textbf{GDPO}), a new policy optimization method to resolve these issues by decoupling the normalization of individual rewards, more faithfully preserving their relative differences and enabling more accurate multi-reward optimization, along with substantially improved training stability. We compare GDPO with GRPO across three tasks: tool calling, math reasoning, and coding reasoning, evaluating both correctness metrics (accuracy, bug ratio) and constraint adherence metrics (format, length). Across all settings, GDPO consistently outperforms GRPO, demonstrating its effectiveness and generalizability for multi-reward reinforcement learning optimization.

\end{abstract}

\begin{document}
\maketitle

\noindent\textbf{Implementations:
\href{https://github.com/NVlabs/GDPO/tree/main/trl-GDPO}{HF-TRL}, 
\href{https://github.com/NVlabs/GDPO/tree/main/verl-GDPO}{verl}, 
\href{https://github.com/NVlabs/GDPO/tree/main/nemo_rl-GDPO}{Nemo-RL}
}
|
\noindent\textbf{Links:
\href{https://nvlabs.github.io/GDPO/}{Project},
\href{https://nv-dler.github.io/}{Lab}
}

\begin{figure}[h]
\begin{center}
\centering
\begin{subfigure}{0.63\textwidth}
  \centering
  \includegraphics[width=\linewidth]{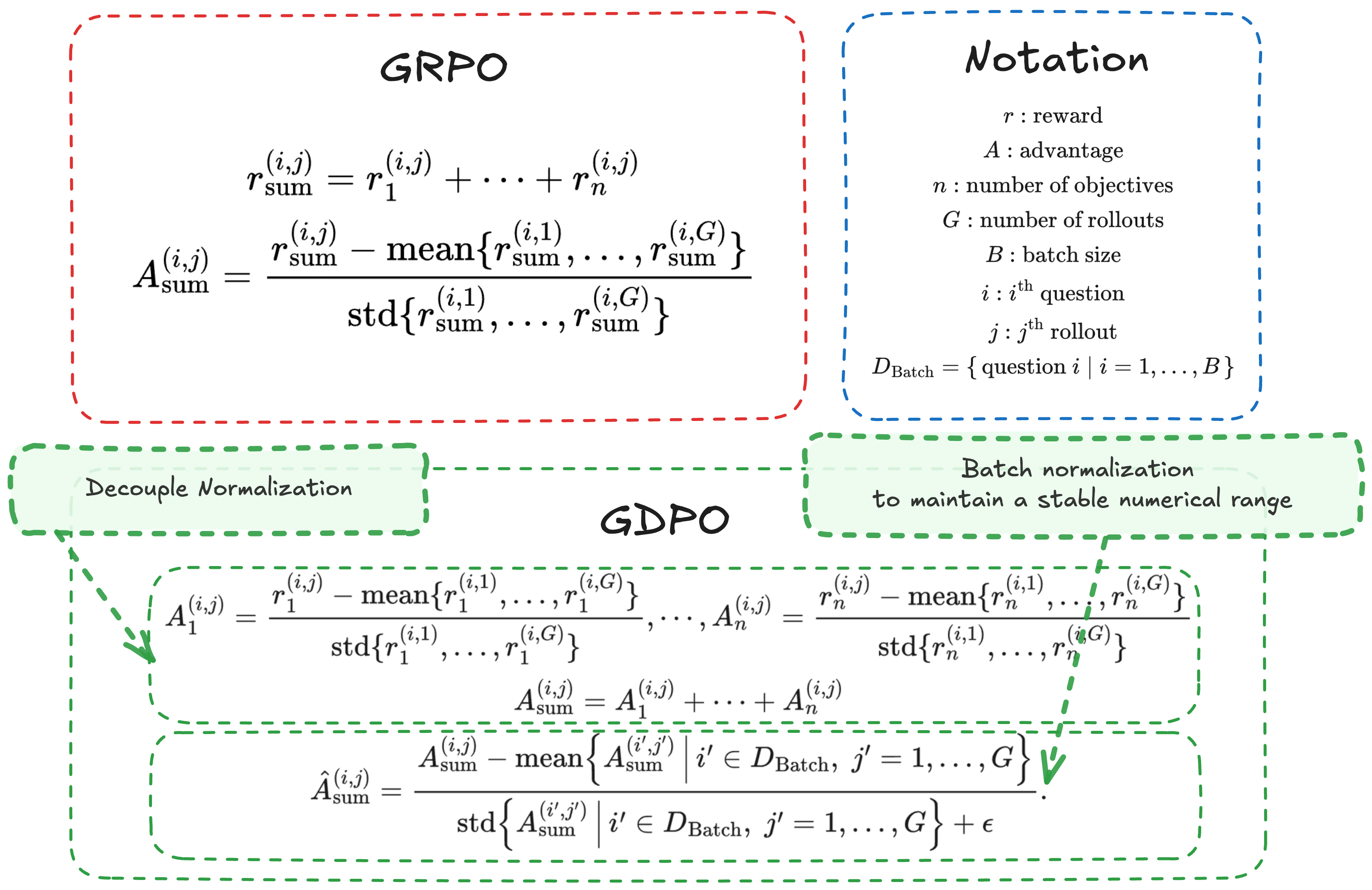}
  \caption{An overview of our proposed GDPO}
  \label{fig:teaser_formulation}
\end{subfigure}
\hfill
\begin{subfigure}{0.33\textwidth}
  \centering
  \vspace{0pt}
  \includegraphics[width=\linewidth]{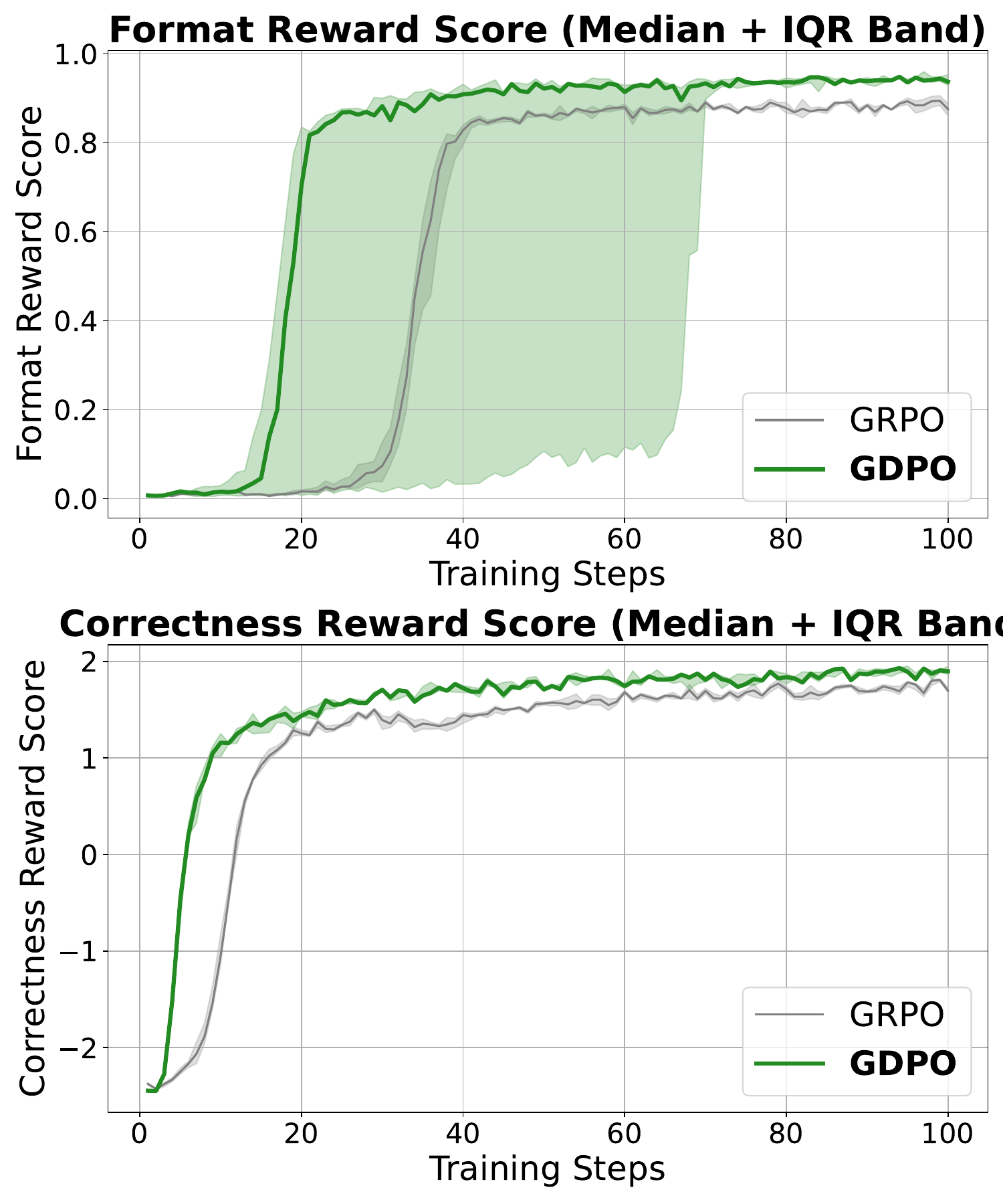}
  \caption{Reward trends: GDPO vs. GRPO}
  \label{fig:teaser_training_curve}
\end{subfigure}
\end{center}
\vspace{-10pt}
\caption{(a): An overview of GDPO, which performs group-wise normalization per reward and then applies batch-wise advantage normalization to preserve a stable numerical range independent of reward count and improve update stability. (b): Median and IQR reward curves over five runs of Qwen2.5-Instruct-1.5B tool-calling RL, demonstrating that GDPO consistently converges to higher correctness and format reward score than GRPO.} 
\end{figure}

\section{Introduction}

\input{sections/intro.tex}

% \section{Preliminaries}
% \input{sections/preliminary.tex}
% \section{GRPO's collapse in multi-reward RL}
\section{GRPO's propensity for reward signal collapse in multi-reward RL}
% \section{GRPO Inherently Compresses Reward Signals in Multi-Reward RL}
\input{sections/motivation.tex}

\section{Method}

\input{sections/method.tex}

\section{Experiments} 
\label{sec:exp}
We begin by evaluating the effectiveness of GDPO compared with GRPO on the tool-calling task (Sec.~\ref{sec:tool_calling}), which involves optimizing two rewards: tool-calling correctness and format compliance. We then present an ablation study that examines the training convergence and downstream performance of GRPO with and without standard deviation normalization. Next, we compare GDPO and GRPO on a math reasoning task that optimizes two implicitly competing rewards, accuracy and length constraint (Sec.~\ref{sec:math_reasoning}). We further conduct extensive analyses of the impact of incorporating different reward weights and modifying reward functions to better reflect varying priorities in human preferences, especially when rewards differ substantially in difficulty. Finally, we extend the number of optimized rewards to three and compare GRPO and GDPO on coding reasoning  (Sec.~\ref{sec:coding_reasoning}), jointly optimizing code-generation accuracy, adherence to length constraints, and bug ratio, further demonstrating that GDPO generalizes effectively to settings with three reward objectives.

\input{sections/exp.tex}

\section{Related Work}
\paragraph{GRPO Variants} 
Several extensions of Group Relative Policy Optimization (GRPO) \cite{Shao2024DeepSeekMathPT} have been proposed to enhance the stability, effectiveness, and efficiency of the framework. These methods explore alternative formulations of group-wise normalization or policy updates while remaining grounded in the core GRPO principles.
For example, to improve stability, Group Sequence Policy Optimization (GSPO) \cite{Zheng2025GroupSP} defines the importance ratio based on sequence likelihood rather than at the token level, performing sequence-level clipping, rewarding, and optimization. To improve RL performance, Decoupled Clip and Dynamic sAmpling Policy Optimization (DAPO) \cite{Yu2025DAPOAO} introduces four key techniques: Clip-Higher, Dynamic Sampling, Token-Level Policy Gradient Loss, and Overlong Reward Shaping.
To promote efficient reasoning, Group Filtered Policy Optimization (GFPO) \cite{Shrivastava2025SampleMT} addresses length explosion by sampling larger groups per problem during training and filtering responses based on their length and reward-per-token ratio. Along the same direction, Doing Length pEnalty Right (DLER) \cite{Liu2025DLERDL} proposes a training recipe combining batch-wise reward normalization, higher clipping, dynamic sampling, and a simple truncation length penalty, achieving state-of-the-art accuracy–efficiency trade-offs.

\paragraph{Multi-Reward Reinforcement Learning}
A growing body of work investigates RL approaches that incorporate multiple reward signals. One major usage is to model diverse human preferences. For example, Safe Reinforcement Learning from Human Feedback \cite{Dai2023SafeRS} decouples human preferences regarding helpfulness and harmlessness, dynamically adjusting the balance between the two objectives during fine-tuning. Similarly, Reinforcement Learning from Personalized Human Feedback (RLPHF) \cite{Jang2023PersonalizedSP} optimizes LLMs for multiple (sometimes conflicting) preferences by training distinct policy models for each preference and merging them during inference. ALARM (Align Language Models via Hierarchical Rewards) \cite{Lai2024ALaRMAL} introduces a hierarchical reward structure that jointly captures dimensions such as response quality, style, fairness, and coherence.
Recent developments in LLMs also integrate multiple-reward optimization to handle complex tasks. For instance, DeepSeek V3.2 \cite{DeepSeekAI2025DeepSeekV32PT} integrates rule-based outcome rewards, length penalties, and language-consistency rewards to enhance reasoning and agentic capabilities.
Another important recent application for multi-reward RL is improving the efficiency of reasoning models while maintaining task performance, primarily by introducing length-based reward functions alongside outcome-based rewards.
For example, O1-Pruner \cite{Luo2025O1PrunerLF} and \cite{Arora2025TrainingLM} apply normalized length penalties to ensure proportional compression. Similarly, \cite{Yi2025ShorterBetterGR} promotes conciseness by penalizing deviations from the shortest correct response within a sampled group. L1 \cite{Aggarwal2025L1CH} introduces Length Controlled Policy Optimization (LCPO) to optimize for accuracy while ensuring responses do not exceed a target length. Finally, \cite{Su2025ThinkingFA} proposes an adaptive reward-shaping method that dynamically adjusts the trade-off between accuracy and response length based on model performance.

%Quick draft:

%GRPO Variants. Several extensions of Group Relative Policy Optimization have been proposed, including GSPO, GFPO, and DAPO, among others. These methods explore alternative formulations of group-wise normalization or policy updates but remain grounded in the GRPO framework.

%Multi-Reward Reinforcement Learning. A growing body of work investigates RL approaches that incorporate multiple reward signals to model diverse human preferences. Safe Reinforcement Learning from Human Feedback focuses on balancing helpfulness with safety-related objectives. DeepSeek V3.2 integrates rule-based outcome rewards, length penalties, and language-consistency rewards. Some other works such as L1, Laser, and KIMI works on improving the efficiency of reasoning model through introducing different length reward function into the optimization. ALARM (Align Language Models via Hierarchical Rewards, 2024) introduces a hierarchical reward structure that jointly captures dimensions such as response quality, style, fairness, and coherence.

%Concurrent Work. MO-GRPO represents a concurrent effort toward multi-reward optimization. However, its evaluation is limited to translation tasks, and its method is effectively equivalent to GDPO without the final batch-normalization stage. Our ablation studies demonstrate that this additional normalization step is crucial for achieving the improved stability and performance observed in our experiments.

\section{Conclusion}

In contrast to prior work that focuses on designing new reward functions for multi-reward reinforcement learning while assuming GRPO is the default optimization method, this study revisits a fundamental but often overlooked question: whether GRPO is actually suitable for multi-reward optimization.
Our analysis shows that applying GRPO directly to the summed reward can cause different reward combinations to collapse into the same advantage values. This collapse eliminates important distinctions across reward dimensions, produces inaccurate policy updates and weaker optimization performance, and can in many cases lead to early training failure. 

To address this limitation, we introduce Group-wise Decoupled Policy Optimization (GDPO), a simple and effective modification to GRPO tailored for multi-reward reinforcement learning. GDPO performs normalization separately for each reward to preserve cross-reward differences, and it incorporates batch-wise advantage normalization to maintain a stable numerical range as additional rewards are included. These changes result in better convergence behavior and models that more faithfully reflect the intended preference structure.

We further present a systematic study for incorporating human preference priorities into the training process and explain how reward functions can be adjusted when the difficulty disparity between objectives is large. Through extensive experiments on tool calling, math reasoning, and coding reasoning, we show that GDPO consistently outperforms GRPO.
Its advantages hold across different numbers of rewards, across different models, and across different reward functions.

Overall, our findings establish GDPO as a more stable, accurate, and preference-aligned optimization method than GRPO for multi-reward reinforcement learning, making it a strong foundation for aligning language models with diverse human preferences in real-world settings.

\newpage
{
  \small
  \bibliographystyle{unsrt}
  \bibliography{main_tech_report}
}

\newpage
\appendix

\section{Training stability issue of GDPO without batch-wise advantage normalization}
\label{sec:gdpo_wo_bn_fail_and_success}
\begin{figure}[h]
\begin{center}
\includegraphics[width=\textwidth]{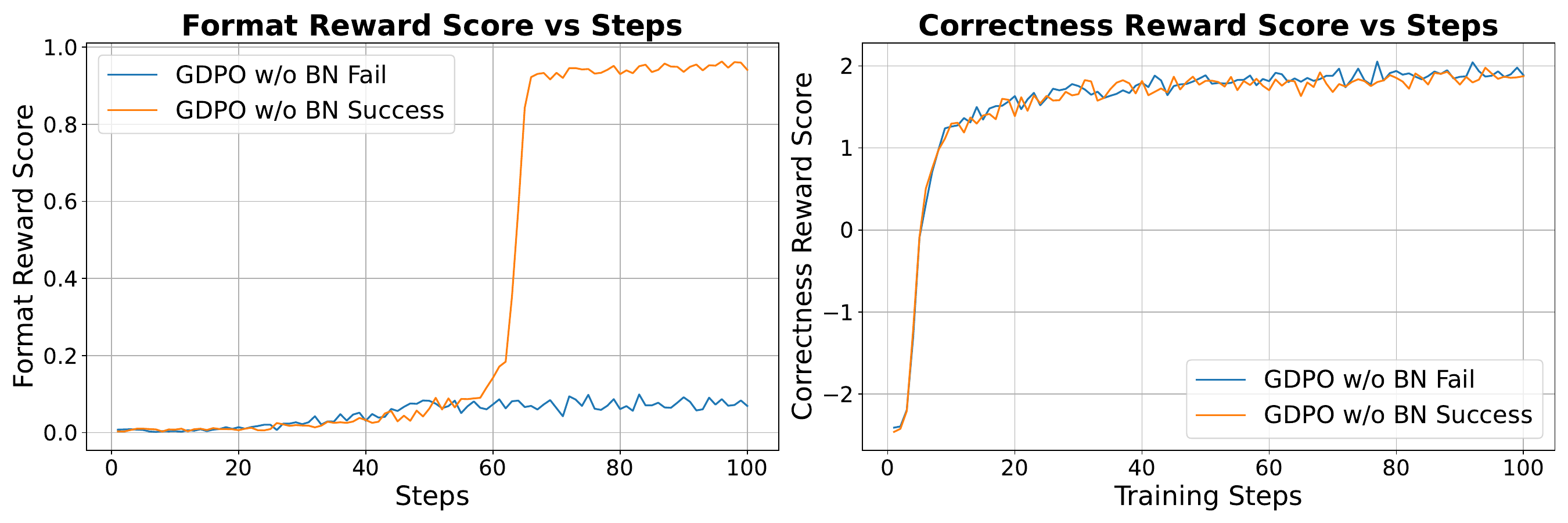}
\end{center}
\caption{Training stability of GDPO with and without batch-wise advantage normalization. Runs without normalization occasionally fail to converge.}
\label{fig:gdpo_fail_success}
\end{figure}

\newpage
\section{ToolRL Training Prompt Format}
\label{sec:toolrl_train_format}
\begin{tcolorbox}[
    colback=blue!80!black,
    colframe=blue!80!black,
    sharp corners,
    boxrule=0pt,
    enlarge left by=-2mm,
    enlarge right by=-2mm,
    enlarge top by=-2mm,
    left=4mm, right=4mm, top=2mm, bottom=2mm
]
{\small\bfseries\color{white} System Prompt for ToolRL Training}
\end{tcolorbox}

\begin{tcolorbox}[
    colback=yellow!10,
    colframe=black!5,
    sharp corners,
    boxrule=0.3pt,
    left=3mm,
    right=3mm,
    top=2mm,
    bottom=2mm
]

You are a helpful dialogue assistant capable of leveraging tool calls to solve user tasks and provide
structured chat responses.

\vspace{2mm}

{\bf Available Tools}\\
In your response, you can use the following tools:\\
\texttt{\{ \{ Tool List \} \}}

\vspace{2mm}

{\bf Steps for Each Turn}
\begin{enumerate}[leftmargin=6mm]
    \item \textbf{Think}: Recall relevant context and analyze the current user goal.
    \item \textbf{Decide on Tool Usage}: If a tool is needed, specify the tool and its parameters.
    \item \textbf{Respond Appropriately}: If a response is needed, generate one while maintaining consistency across user queries.
\end{enumerate}

\vspace{2mm}

{\bf Output Format}

\textcolor{thinkblue}{\texttt{\textless think\textgreater}} Your thoughts and reasoning \textcolor{thinkblue}{\texttt{\textless/think\textgreater}}

\textcolor{toolgreen}{\texttt{\textless tool\_call\textgreater}}
\{``name'': ``Tool name'', ``parameters'': \{``Parameter name'': ``Parameter content'', `` ... ...'': `` ... ...''\}\}\\
\{``name'': `` ... ...'', ``parameters'': \{`` ... ...'': `` ... ...'', `` ... ...'': `` ... ...''\}\}\\
...\\
\textcolor{toolgreen}{\texttt{\textless /tool\_call\textgreater}}

\textcolor{responsepurple}{\texttt{\textless response\textgreater}}AI's final response \textcolor{responsepurple}{\texttt{\textless /response\textgreater}}

\vspace{2mm}

{\bf Important Notes}
\begin{enumerate}[leftmargin=6mm]
    \item You must always include the \texttt{\textless think\textgreater} field to outline your reasoning. Provide at least one of \texttt{\textless tool\_call\textgreater} or \texttt{\textless response\textgreater}. Decide whether to use \texttt{\textless tool\_call\textgreater} (possibly multiple times), \texttt{\textless response\textgreater}, or both.
    \item You can invoke multiple tool calls simultaneously in the \texttt{\textless tool\_call\textgreater} fields. Each tool call should be a JSON object with a “name'' field and a “parameters'' field containing a dictionary of parameters. If no parameters are needed, leave the “parameters'' field an empty dictionary.
    \item Refer to the previous dialogue records in the history, including the user’s queries, previous \texttt{\textless tool\_call\textgreater}, \texttt{\textless response\textgreater}, and any tool feedback noted as \texttt{\textless obs\textgreater} (if exists).
\end{enumerate}

\end{tcolorbox}

% \begin{tcolorbox}[
%     enhanced,
%     colback=boxbg,
%     colframe=boxborder,
%     width=\linewidth,
%     arc=12pt,
%     boxrule=1pt,
%     top=0pt,
%     left=6pt,
%     right=6pt,
%     bottom=6pt,
%     enlarge top by=0pt,
%     overlay={
%         % HEADER BAR
%         \fill[headerblue] 
%             ([xshift=-1pt,yshift=11pt]frame.north west) 
%             rectangle 
%             ([xshift=1pt,yshift=-20pt]frame.north east);
%         % HEADER TEXT
%         \node[anchor=west, yshift=-7pt, xshift=8pt, text=white, font=\large\bfseries]
%             at ([yshift=0pt]frame.north west)
%             {User Prompt for Training};
%     }
% ]
\begin{tcolorbox}[
    colback=blue!80!black,
    colframe=blue!80!black,
    sharp corners,
    boxrule=0pt,
    enlarge left by=-2mm,
    enlarge right by=-2mm,
    enlarge top by=-2mm,
    left=4mm, right=4mm, top=2mm, bottom=2mm
]
{\small\bfseries\color{white} User Prompt for ToolRL Training}
\end{tcolorbox}
\begin{tcolorbox}[
    colback=yellow!10,
    colframe=black!5,
    sharp corners,
    boxrule=0.3pt,
    left=3mm,
    right=3mm,
    top=2mm,
    bottom=2mm
]
\vspace{18pt} % space below header banner

{\bf Dialogue History}\\

\texttt{\textless user\textgreater} \{ \{ Initial User Input \} \} \texttt{\textless/user\textgreater}

\vspace{8pt}

\textcolor{thinkblue}{\texttt{\textless think\textgreater}}
\; Round 1 Model Thought \;
\textcolor{thinkblue}{\texttt{\textless/think\textgreater}}

\{ \{ Round 1 model output 
\textcolor{toolgreen}{\texttt{\textless tool\_call\textgreater}}
or 
\textcolor{responsepurple}{\texttt{\textless response\textgreater}}
\} \}

\textcolor{obsbrown}{\texttt{\textless obs\textgreater}} 
Round 1 Observation 
\textcolor{obsbrown}{\texttt{\textless/obs\textgreater}}

\vspace{10pt}
... ...

\vspace{12pt}

\texttt{\textless user\textgreater} \{ \{ User Input \} \} \texttt{\textless/user\textgreater}

\vspace{8pt}
... ...

\end{tcolorbox}

\newpage
\section{Tool Calling Reward Functions}
\label{sec:tool_calling_reward_function}
\paragraph{Format Reward.}
The format reward $\mathcal{R}_{\text{format}} \in \{0,1\}$ checks whether the model output satisfies the required structure and contains all necessary fields in the correct order:
\begin{equation}
    \mathcal{R}_{\text{format}} =
    \begin{cases}
    1, & \text{if all required fields appear and are in the correct order}, \\
    0, & \text{otherwise}.
    \end{cases}
\end{equation}

\paragraph{Correctness Reward.}
The correctness reward $\mathcal{R}_{\text{correct}} \in [-3,\,3]$ evaluates the predicted tool calls 
$P = \{P_1, \ldots, P_m\}$ against the ground-truth calls 
$G = \{G_1, \ldots, G_n\}$. It consists of three components:

\begin{itemize}

\item \textbf{Tool Name Matching:}\\[2mm]
\[
r_{\text{name}} = 
\frac{|N_G \cap N_P|}{|N_G \cup N_P|}
\in [0,1],
\]
where $N_G$ and $N_P$ are the sets of tool names from ground-truth and predicted calls, respectively.

\item \textbf{Parameter Name Matching:}\\[2mm]
\[
r_{\text{param}} = 
\sum_{G_j \in G}
\frac{|\mathrm{keys}(G_j) \cap \mathrm{keys}(P_j)|}{
      |\mathrm{keys}(G_j) \cup \mathrm{keys}(P_j)|}
\in [0, |G|],
\]
where $\mathrm{keys}(G_j)$ and $\mathrm{keys}(P_j)$ are the parameter names of the ground-truth and predicted calls.

\item \textbf{Parameter Content Matching:}\\[2mm]
\[
r_{\text{value}} =
\sum_{G_j \in G}
\sum_{k \in \mathrm{keys}(G_j)}
\mathbf{1}[P_G[k] = P_P[k]]
\in \left[0,\; \sum_{G_j \in G} |\mathrm{keys}(G_j)| \right],
\]
where $P_G[k]$ and $P_P[k]$ are the parameter values for the ground-truth and predicted calls.

\item \textbf{Total Match Score:}\\[2mm]
\[
r_{\text{match}} = 
r_{\text{name}} + r_{\text{param}} + r_{\text{value}}
\in [0, S_{\max}],
\]
where
\[
S_{\max} = 1 + |G| + \sum_{G_j \in G} |\mathrm{keys}(G_j)|.
\]

\end{itemize}

The final correctness reward is computed by finding the optimal matching between $P$ and $G$ to maximize the total match score:
\[
\mathcal{R}_{\text{correct}}
= 6 \cdot \frac{R_{\max}}{S_{\max}} - 3
\in [-3,\,3].
\]
where $R_{\max}$ denotes the total match score from the optimal matching.

\newpage
\section{ToolRL Hyperparameters Setting}
\label{sec:toolrl_hyperparameters}

\begin{table}[!ht]
\centering
\caption{GDPO verl training configuration. All hyperparameter settings are kept identical to those used in ToolRL\cite{qian2025toolrl}.}
\begin{tabular}{l l}
\toprule
\textbf{Parameter} & \textbf{Value} \\
\midrule
trainer.total\_epochs & 15 \\
data.train\_batch\_size & 512 \\
actor\_rollout\_ref.actor.ppo\_mini\_batch\_size & 128 \\
data.max\_prompt\_length & 2048 \\
actor\_rollout\_ref.actor.optim.lr & 1.00E-06 \\
actor\_rollout\_ref.rollout.n & 4 \\
algorithm.kl\_ctrl.kl\_coef & 0.001 \\
\bottomrule
\end{tabular}
\end{table}

\newpage
\section{Math/Coding Reasoning Hyperparameters Setting}
\label{sec:math_code_reasoning_hyperparameters}

\begin{table}[!ht]
\centering
\caption{GDPO verl training configuration}
\begin{tabular}{l l}
\toprule
\textbf{Parameter} & \textbf{Value} \\
\midrule
data.train\_batch\_size & 512 \\
actor\_rollout\_ref.actor.ppo\_mini\_batch\_size & 64 \\
actor\_rollout\_ref.actor.ppo\_epochs & 1 \\
data.max\_prompt\_length & 1024 \\
actor\_rollout\_ref.actor.optim.lr & 1.00E-06 \\
actor\_rollout\_ref.rollout.temperature & 1 \\
actor\_rollout\_ref.rollout.n & 16 \\
actor\_rollout\_ref.actor.clip\_ratio\_low & 0.2 \\
actor\_rollout\_ref.actor.clip\_ratio\_high & 0.28 \\
algorithm.filter\_groups.enable & TRUE \\
algorithm.filter\_groups.metric & seq\_reward \\
actor\_rollout\_ref.actor.kl\_loss\_coef & 0.0005 \\
actor\_rollout\_ref.actor.kl\_loss\_type & mse \\
\bottomrule
\end{tabular}
\end{table}

\newpage
\section{Training curves of GRPO and GDPO when training DeepSeek-R1-7B and Qwen3-4B-Instruct with $\mathcal{R}_{\text{length}}$ and $\mathcal{R}_{\text{correct}}$ on math reasoning data.}
\begin{figure}[h]
\begin{center}
\includegraphics[width=\textwidth]{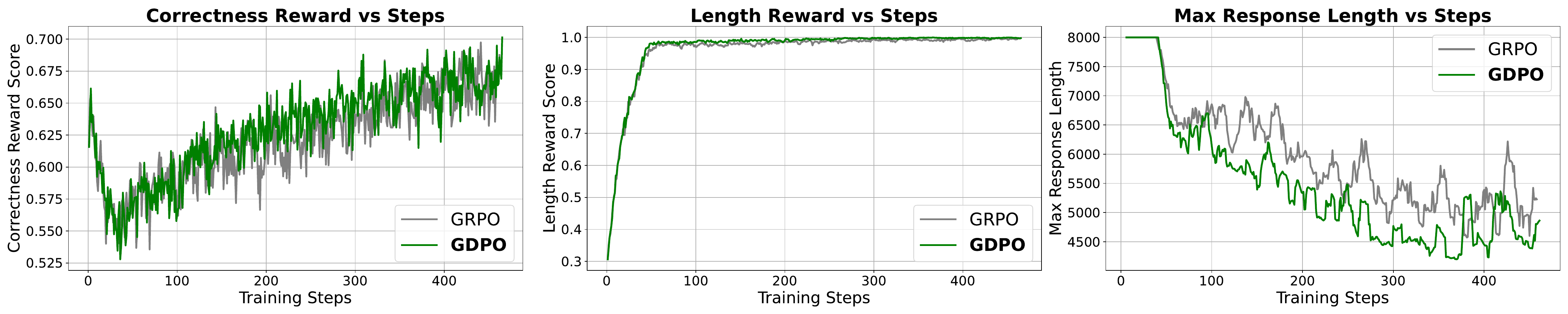}
\end{center}
\caption{Training behavior of GRPO and GDPO when optimizing DeepSeek-R1-7B across correctness reward, length reward, and maximum batch response length on math reasoning data. We can see that GDPO maintains improving correctness and better adherence to length constraints over GRPO.}
\label{fig:math_7b_training_curve}
\end{figure}

\begin{figure}[h]
\begin{center}
\includegraphics[width=\textwidth]{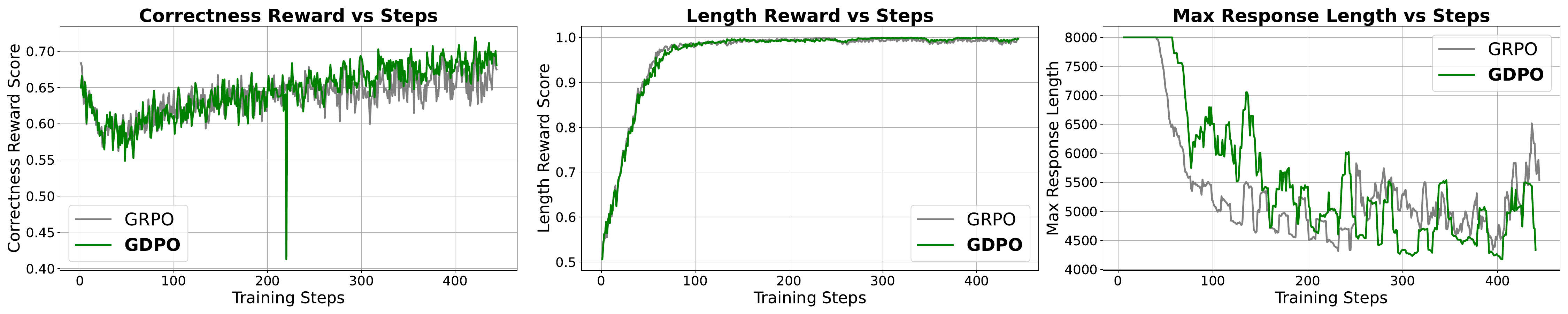}
\end{center}
\caption{Training behavior of GRPO and GDPO when optimizing Qwen3-4B-Instruct across correctness reward, length reward, and maximum batch response length on math reasoning data. We can see that GDPO maintains improving correctness and better adherence to length constraints over GRPO.}
\label{fig:math_4b_training_curve}
\end{figure}

\newpage
\section{Comparison of GRPO/GDPO finetuned DeepSeek-R1-7B models under varying length reward weights $\{1.0, 0.75, 0.5, 0.25\}$ with and without the conditioned length reward $\tilde{\mathcal{R}}_{\text{length}}$ on math reasoning tasks}
\label{sec:diff_weight_appendix}

\begin{table}[!htp]\centering
\caption{Comparison of GRPO/GDPO finetuned DeepSeek-R1-7B models under varying length reward weights $\{1.0, 0.75, 0.5, 0.25\}$ with the normal length reward $\mathcal{R}_{\text{length}}$ on math reasoning tasks}\label{tab: }
\resizebox{\textwidth}{!}{ % use this if the table is too large
\begin{tabular}{lcrrrrrrrrrr|rrr}\toprule
&Length Reward Weight &MATH ↑ &Exceed ↓ &AIME ↑ &Exceed ↓ &Amc ↑ &Exceed ↓ &Minerva ↑ &Exceed ↓ &Olympiad ↑ &Exceed ↓ & \textbf{Avg Acc ↑} & \textbf{Avg Exceed ↓} \\\cmidrule{2-14}
DeepSeek-R1-7B &- &93.6\% &26.0\% &55.4\% &85.6\% &82.9\% &57.2\% &49.8\% &41.8\% &58.2\% &60.6\% &68.0\% &54.2\% \\\cmidrule{1-14}
\multirow{4}{*}{GRPO-$\mathcal{R}_{\text{length}}$} &0.25 &94.0\% &0.5\% &53.3\% &4.9\% &83.8\% &0.6\% &53.2\% &1.9\% &59.8\% &2.4\% &68.8\% &2.1\% \\
&0.5 &94.2\% &0.6\% &52.1\% &2.3\% &83.2\% &0.8\% &53.9\% &0.2\% &60.2\% &0.8\% &68.7\% &0.9\% \\
&0.75 &93.5\% &0.3\% &51.2\% &2.5\% &83.0\% &0.5\% &53.4\% &0.1\% &58.2\% &1.5\% &67.8\% &1.0\% \\
&1.0 &94.1\% &0.5\% &50.2\% &2.1\% &83.8\% &0.6\% &53.2\% &0.2\% &60.2\% &1.1\% &68.3\% &0.9\% \\\cmidrule{1-14}
\multirow{4}{*}{GDPO-$\mathcal{R}_{\text{length}}$} &0.25 &94.2\% &0.5\% &54.7\% &3.9\% &84.5\% &1.4\% &54.1\% &1.1\% &59.1\% &1.3\% &69.3\% &1.6\% \\
&0.5 &93.2\% &0.2\% &53.8\% &0.8\% &84.1\% &0.4\% &53.3\% &0.2\% &58.5\% &1.1\% &68.6\% &0.5\% \\
&0.75 &93.6\% &0.2\% &54.4\% &1.5\% &83.4\% &0.2\% &53.4\% &0.3\% &58.8\% &0.6\% &68.7\% &0.5\% \\
&1.0 &93.9\% &0.1\% &53.1\% &0.2\% &84.0\% &0.3\% &53.8\% &0.1\% &59.3\% &0.4\% &68.8\% &0.2\% \\\midrule
\bottomrule
\end{tabular}%
}
\end{table}

\begin{table}[!htp]\centering
\caption{Comparison of GRPO/GDPO finetuned DeepSeek-R1-7B models under varying length reward weights $\{1.0, 0.75, 0.5, 0.25\}$ with the conditioned length reward $\tilde{\mathcal{R}}_{\text{length}}$ on math reasoning tasks}\label{tab: }
\resizebox{\textwidth}{!}{ % use this if the table is too large
\begin{tabular}{lcrrrrrrrrrr|rrr}\toprule
&Length Reward Weight &MATH ↑ &Exceed ↓ &AIME ↑ &Exceed ↓ &Amc ↑ &Exceed ↓ &Minerva ↑ &Exceed ↓ &Olympiad ↑ &Exceed ↓ & \textbf{Avg Acc ↑} & \textbf{Avg Exceed ↓} \\\cmidrule{2-14}
DeepSeek-R1-7B &- &93.6\% &26.0\% &55.4\% &85.6\% &82.9\% &57.2\% &49.8\% &41.8\% &58.2\% &60.6\% &68.0\% &54.2\% \\\cmidrule{1-14}
\multirow{4}{*}{GRPO-$\mathcal{\tilde{R}}_{\text{length}}$} &0.25 &94.4\% &6.4\% &56.8\% &51.2\% &85.8\% &18.2\% &54.5\% &6.5\% &61.2\% &18.8\% &70.6\% &20.2\% \\
&0.5 &94.2\% &4.0\% &56.5\% &43.8\% &85.9\% &13.6\% &54.8\% &7.1\% &60.6\% &17.2\% &70.4\% &17.1\% \\
&0.75 &94.1\% &3.4\% &55.5\% &38.3\% &85.2\% &11.2\% &54.3\% &5.7\% &60.8\% &14.3\% &70.0\% &14.6\% \\
&1.0 &93.2\% &2.7\% &53.3\% &29.2\% &82.9\% &8.6\% &53.2\% &2.5\% &59.1\% &10.3\% &68.3\% &10.7\% \\\cmidrule{1-14}
\multirow{4}{*}{GDPO-$\mathcal{\tilde{R}}_{\text{length}}$} &0.25 &94.9\% &5.1\% &57.7\% &32.8\% &86.2\% &11.2\% &54.9\% &5.1\% &62.1\% &14.6\% &71.2\% &13.8\% \\
&0.5 &94.5\% &3.1\% &57.7\% &32.1\% &85.8\% &9.3\% &54.1\% &3.9\% &61.6\% &13.3\% &70.8\% &12.3\% \\
&0.75 &94.6\% &2.8\% &56.0\% &31.9\% &86.9\% &10.3\% &53.0\% &2.6\% &61.3\% &12.8\% &70.4\% &12.1\% \\
&1.0 &93.9\% &1.4\% &57.7\% &12.3\% &85.9\% &3.8\% &53.4\% &1.0\% &60.8\% &6.2\% &70.4\% &4.9\% \\\midrule
\bottomrule
\end{tabular}%
}
\end{table}

%%%%%%%%%%%%%%%%%%%%%%%%%%%%%%%%%%%%%%%%%%%%%%%%%%%%%%%%%%%%%%%%%%%%%%%%%%%%%%%
%%%%%%%%%%%%%%%%%%%%%%%%%%%%%%%%%%%%%%%%%%%%%%%%%%%%%%%%%%%%%%%%%%%%%%%%%%%%%%%
% APPENDIX
%%%%%%%%%%%%%%%%%%%%%%%%%%%%%%%%%%%%%%%%%%%%%%%%%%%%%%%%%%%%%%%%%%%%%%%%%%%%%%%
%%%%%%%%%%%%%%%%%%%%%%%%%%%%%%%%%%%%%%%%%%%%%%%%%%%%%%%%%%%%%%%%%%%%%%%%%%%%%%%

\end{document}

%% file: math_commands.tex
\def\eqref#1{equation~\ref{#1}}
\def\1{\bm{1}}

\DeclareMathAlphabet{\mathsfit}{\encodingdefault}{\sfdefault}{m}{sl}
\SetMathAlphabet{\mathsfit}{bold}{\encodingdefault}{\sfdefault}{bx}{n}

%% file: sections/intro.tex
As language models continue to advance in capability, expectations for their behavior have grown accordingly. Demand for models to not only provide accurate responses but also exhibit behaviors aligned with a wide range of human preferences across diverse scenarios has continued to increase. These preferences span efficiency~\cite{liu2025laser, team2025kimi, aggarwal2025l1}, safety~\cite{mu2024rule}, response coherence and logic~\cite{chen2025grpocareconsistencyawarereinforcementlearning,liu2025deepseek}, gender biases~\cite{zhang2025genderalign} and many other objectives. Meeting such heterogeneous requirements within a single model is a challenging task.

% \peter{You had both parskip and parindent active; I set parindent to 0pt}

Reinforcement learning (RL) has emerged as the de facto training pipeline for aligning large language models to fulfill such diverse human preferences. In particular, recent RL-based approaches have begun to incorporate multiple rewards into training, with each reward designed to capture different human preferences and collectively guide models toward human-favored behaviors. Despite this growing interest in multi-reward RL, recent work~\cite{liu2025laser,aggarwal2025l1,chen2025grpocareconsistencyawarereinforcementlearning} has largely focused on the reward design itself and often directly relied on applying Group Relative Policy Optimization (GRPO) directly for multi-reward RL optimization, often without examining whether GRPO is well-suited for optimizing combinations of heterogeneous rewards.

In this paper, we revisit the applicability of GRPO in multi-reward settings and show that directly applying GRPO to normalize different combinations of rollout rewards can cause them to collapse into identical advantage values, which effectively limits the precision of the training signal, as illustrated in Fig.~\ref{fig:advantage_illustration_grpo_gdpo}. This collapse removes important distinctions across reward dimensions and leads to inaccurate policy updates, suboptimal reward convergence, and, in many cases, early training failure.

To overcome these challenges, we propose \textbf{Group reward-Decoupled Normalization Policy
Optimization (GDPO)} which decouples the group-wise normalization of each individual reward as illustrated in Fig.~\ref{fig:teaser_formulation},  to ensure that distinctions across different reward combinations are better preserved and more accurately reflect the relative differences in model responses. This leads to more precise multi-reward optimization and substantially improved training convergence. After this decoupled group-wise normalization, we apply batch-wise advantage normalization to ensure that the magnitude of advantage does not increase as the number of individual rewards increases.

We compare GDPO and GRPO across three tasks: tool calling, math reasoning, and code reasoning. These tasks cover a wide range of objectives, including tool-calling accuracy and format correctness, mathematical reasoning accuracy and adherence to reasoning-length constraints, and code pass rate and bug ratio. Across all tasks, GDPO converges better.
For example, in Fig.~\ref{fig:teaser_training_curve}, training Qwen2.5-1.5B-Instruct with GDPO attains both higher correctness and format compliance than GRPO on the tool-calling task. On challenging math tasks, GDPO consistently outperforms GRPO. For instance, training DeepSeek-R1-1.5B and Qwen3-4B-Instruct with GDPO yields up to 6.3\% and 2.3\% higher accuracy on AIME compared to GRPO, while keeping more responses short simultaneously.

Taken together, these results demonstrate the effectiveness and generalizability of GDPO, showing it to be a better alternative to GRPO for multi-reward RL optimization.

Our contributions are as follows:
\begin{itemize}
    \item \textbf{Analysis of GRPO reward collapse.} We demonstrate that applying GRPO naively for multi-reward RL optimization can collapse distinct rollout reward combinations into identical advantage values, thereby diminishing the resolution of the learning signal.

    \item \textbf{Remediation of GRPO reward collapse.} We propose GDPO, which performs group-wise decoupled normalization of each reward separately to better preserve cross-reward distinctions and enable more accurate multi-reward optimization.
    
    \item In addition to GDPO, we provide a systematic overview of how to modify reward functions and adjust reward weights to more faithfully align with preferences of varying priority.

\item We carry out extensive experiments on three tasks: tool calling, math reasoning, and code reasoning, and compare the effectiveness of GDPO on optimizing a wide range of rewards corresponding to accuracy, format correctness, length constraints, and code quality. 
In all settings, GDPO consistently outperforms GRPO, showing improved training convergence and stronger downstream performance that align more closely with a diverse set of preferences.
\end{itemize}

%% file: sections/motivation.tex
Recent advancements such as Group Relative Policy Optimization (GRPO)~\citep{guo2025deepseek} and its variants, including DAPO~\citep{yu2025dapo} and Reinforce++-Baseline~\citep{hu2025reinforce++}, have emerged as widely adopted reinforcement learning algorithms due to their efficiency and simplicity. 
In contrast to Proximal Policy Optimization (PPO)~\cite{schulman2017proximal}, GRPO eliminates the need for a value model by leveraging group-relative advantage estimation for policy updates.

Currently, GRPO has been primarily employed for optimizing a single-objective reward, typically focusing on accuracy. 
However, as model capability continues to grow, recent works have increasingly sought to optimize multiple rewards, such as response length constraint and formatting quality, in addition to accuracy~\citep{liu2025laser,qian2025toolrl,aggarwal2025l1}, to better align with human preferences. Existing approaches for multi-reward RL generally adopt a straightforward strategy: summing all reward components and applying GRPO directly.

Formally, for a given question–answer pair $(q_i, o_j)$, where the behavior policy $\pi_{\theta_{\mathrm{old}}}$ samples a group of $G$ responses $\{o_j\}_{j=1}^G$, and assuming $n$ objectives, the aggregated reward for the $j$-th response is computed as the sum of each objective’s reward:
\begin{equation}
r^{(i,j)}_{\text{sum}} = r_1^{(i,j)} + \cdots + r_n^{(i,j)}
\label{eq:grpo_sum}
\end{equation}
The group-relative advantage for the $j$-th response is then obtained by normalizing the group-level aggregated rewards:
\begin{equation}
A^{(i,j)}_{\text{sum}} =
\frac{
r_{\text{sum}}^{(i,j)} - \mathrm{mean}\{ r_{\text{sum}}^{(i,1)}, \ldots, r_{\text{sum}}^{(i,G)}\}
}{
\mathrm{std}\{ r_{\text{sum}}^{(i,1)}, \ldots, r_{\text{sum}}^{(i,G)} \}
}
\label{eq:grpo_advantage}
\end{equation}
The corresponding multi-reward GRPO optimization objective can then be expressed as:

\begin{equation}
\mathcal{J}_{\mathrm{GRPO}}(\theta) = 
\mathbb{E}_{(q_i,o_j) \sim D,\; \{o_j\}_{j=1}^G \sim \pi_{\theta_{\mathrm{old}}}(\cdot|q)} 
\left[
\frac{1}{G} \sum_{j=1}^{G} \frac{1}{|o_j|} \sum_{t=1}^{|o_j|}
\min \left(
s_{i,t}(\theta)\, A^{(i,j)}_{\text{sum}},\;
\mathrm{clip}(s_{i,t}(\theta), 1 - \epsilon, 1 + \epsilon)\,A^{(i,j)}_{\text{sum}}
\right)
\right]
\label{eq:grpo}
\end{equation}
where $s_t(\theta) = \frac{\pi_{\theta}(o_j^t \,|\, q, o_j^{<t})}{\pi_{\theta_{\mathrm{old}}}(o_j^t \,|\, q, o_j^{<t})}$ 
and $\epsilon$ denotes the clipping threshold. For clarity, we omit the KL-divergence loss term in this formulation. 

\begin{figure}[h]
\begin{center}
\includegraphics[width=\textwidth]{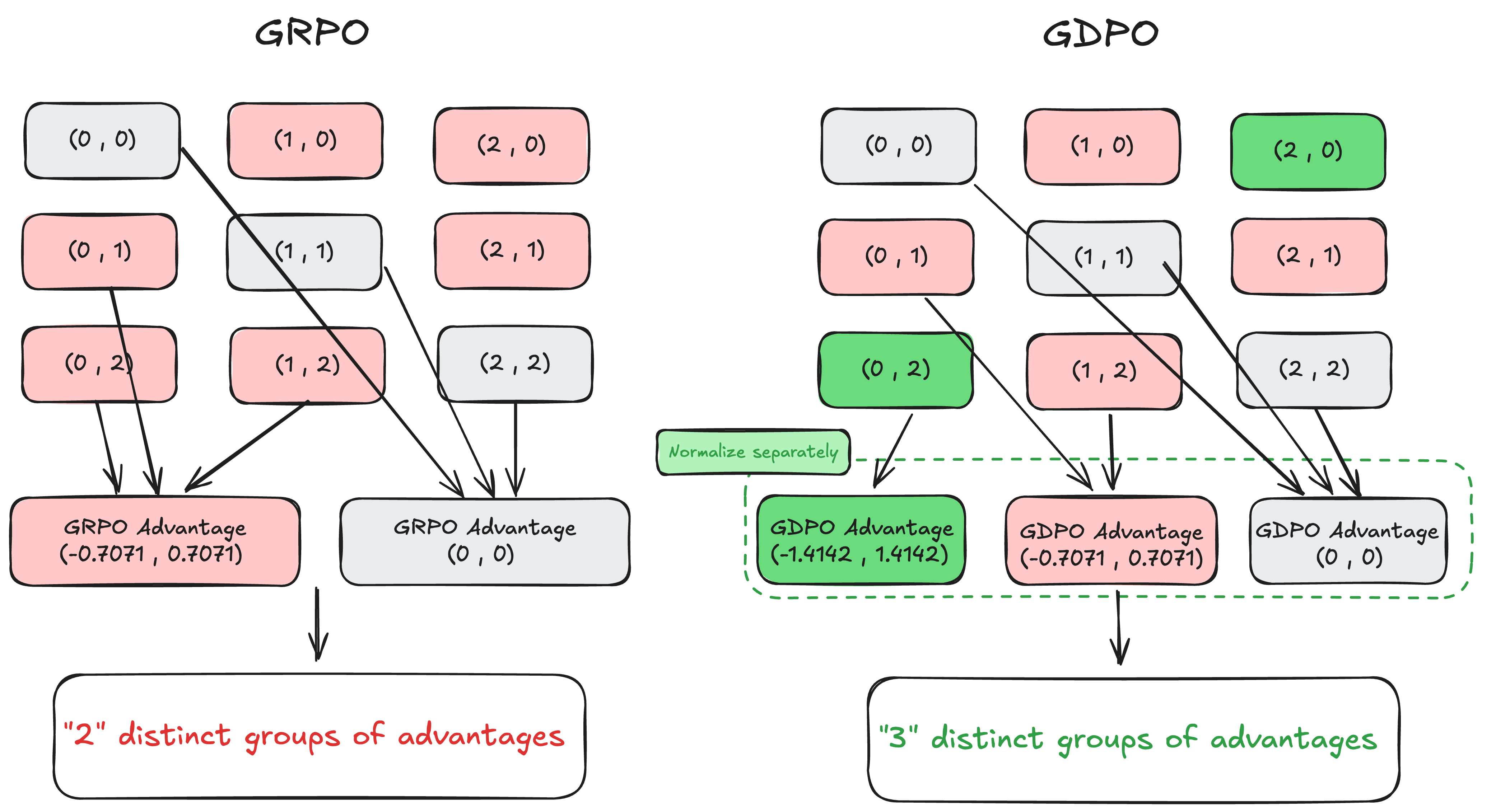}
\end{center}
\caption{Comparison of GRPO and GDPO advantage computation in a two-binary-reward, two-rollout example. GRPO maps different reward combinations into only two distinct advantage groups, whereas GDPO normalizes each reward independently and retains three distinct groups of advantage values. We skip the batch-wise normalization calculation step in GDPO here for simplicity since it does not change the number of distinct advantage groups.}
\label{fig:advantage_illustration_grpo_gdpo}
\end{figure}

\begin{figure}[h]
\begin{center}
\includegraphics[width=\textwidth]{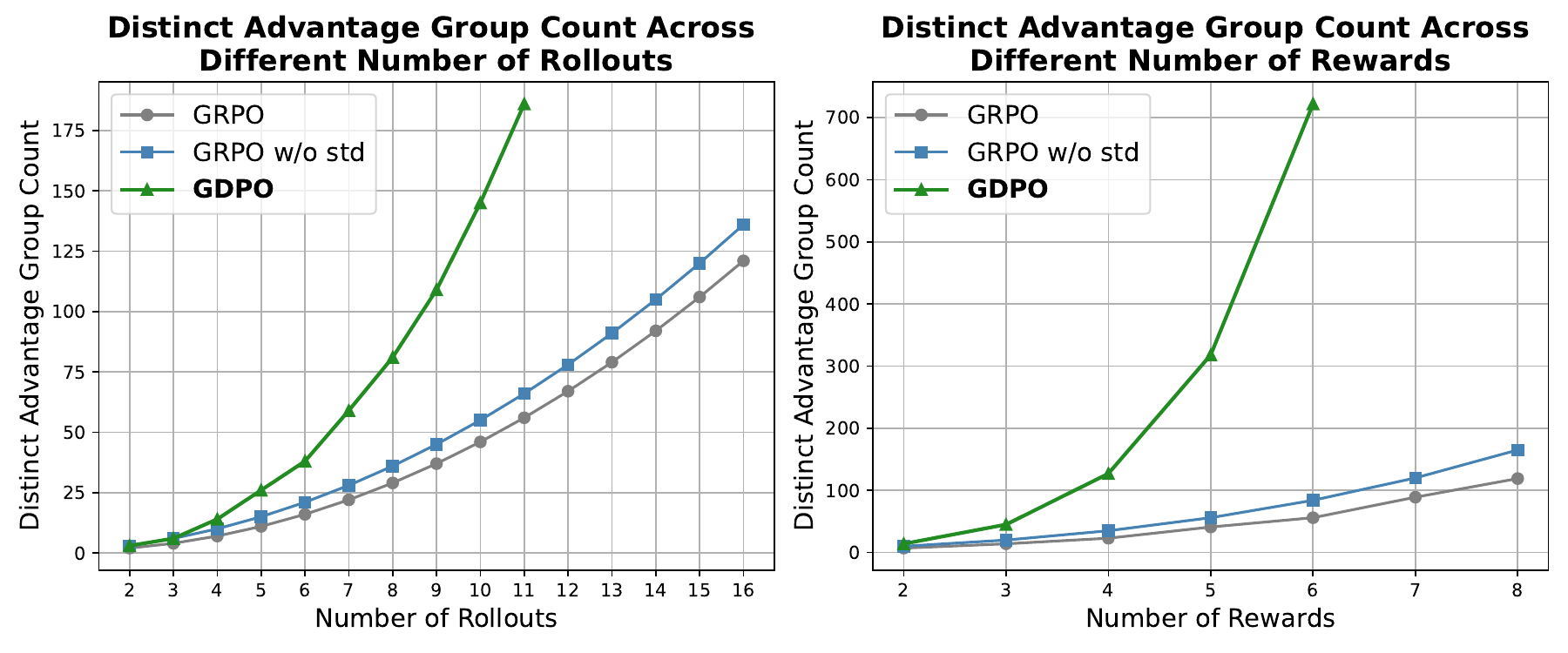}
\end{center}
\caption{Comparison of the number of distinct advantage groups produced by GRPO, GRPO without standard deviation normalization (GRPO w/o std), and GDPO. As the number of rollouts (left) or rewards (right) grows, GDPO consistently preserve a substantially larger number of distinct advantage groups compared to GRPO and GRPO w/o std. This results in advantage estimations that provide more expressive training signals.}
\label{fig:advantage_count_plot}
\end{figure}

We first revisit this common practice of applying GRPO for mulit-reward RL optimization and identify a previously overlooked issue, that is GRPO inherently compresses the reward signal, causing loss of information in the advantage estimates. To illustrate, we start with a simple training setting and then extend it to more general cases. Consider a scenario where we generate two rollouts for each question for calculating the group-relative advantage and the task involves two binary reward $r_1, r_2 \in \{0,1\}$. Consequently, the total reward for each rollout can take values from $\{0,1,2\}$.

As shown in Figure~\ref{fig:advantage_illustration_grpo_gdpo}, we enumerate all possible rollout reward combinations within a group, represented as $(\mathtt{rollout}\text{ } \mathtt{1}'\mathtt{s}\text{ } \mathtt{total}\text{ } \mathtt{reward}, \mathtt{rollout}\text{ } \mathtt{2's}\text{ } \mathtt{total}\text{ } \mathtt{reward})$ and the corresponding normalized advantages as $(\mathit{rollout}\text{ } \mathit{1}'\mathit{s}\text{ } \mathit{normalized}\text{ } \mathit{advantage}, \mathit{rollout}\text{ } \mathit{2's}\text{ } \mathit{normalized}\text{ } \mathit{advantage})$. Despite having six distinct combinations when order is ignored, only two unique advantage groups emerge after applying group-wise reward normalization. Specifically, $(\mathtt{0,1})$, $(\mathtt{0,2})$, and $(\mathtt{1,2})$ yield identical normalized advantages $A_{\text{sum}}$ of $(\mathit{-0.7071},\, \mathit{0.7071})$, while $(\mathtt{0,0})$, $(\mathtt{1,1})$, and $(\mathtt{2,2})$ all result in $(\mathit{0,0})$.

This demonstrates a fundamental limitation of GRPO’s advantage calculation in multi-reward optimization which over-compresses the rich group-wise reward signal. Intuitively, $\mathtt{(0,2)}$ should produce a stronger learning signal than $\mathtt{(0,1)}$ because a total reward of $2$ indicates simultaneous satisfaction of two rewards, whereas a reward of $1$ corresponds to achieving only one. Thus, when the other one rollout only receives zero reward, $\mathtt{(0,2)}$ should yield a larger relative advantage than $\mathtt{(0,1)}$. This limitation can also introduce risks of training instability due to inaccurate advantage estimates. As shown in Fig.~\ref{fig:math_1.5b_training_curve}, the correctness reward score begins to decline after approximately 400 training steps when training with GRPO, indicating a partial training collapse.

Recently, Dr.GRPO~\cite{liu2025understanding} and DeepSeek-v3.2~\cite{liu2025deepseek} adopt a variant of GRPO that removes the standard deviation normalization term from Eq.~\ref{eq:grpo_advantage}, such that $A^{(i,j)}_{\text{sum}} = r_{\text{sum}}^{(i,j)} - \mathrm{mean}\{r_{\text{sum}}^{(i,1)}, \ldots, r_{\text{sum}}^{(i,G)} \}$.
Despite these works introduce this modification to mitigate question-level difficulty bias, at first glance, this change also appears to address the issue we identify. Specifically, removing the standard deviation normalization mitigates the issue: $\mathtt{(0,1)}$ and $\mathtt{(0,2)}$ now yield distinct advantages of $\mathit{(-0.5, 0.5)}$ and $\mathit{(-1.0, 1.0)}$, respectively. However, when this setup is generalized to a larger number of rollouts while keeping the number of rewards fixed, as shown in Figure~\ref{fig:advantage_count_plot}, we observe that such fix only slightly increases the number of distinct advantage groups compared to GRPO. A similar trend can be observed under settings where the number of rollouts is fixed at four, but the number of rewards gradually increases. In this case, we also observe only modest improvements in the number of distinct advantage groups. We also empirically examine the effectiveness of removing the standard deviation normalization term in Section~\ref{sec:different_component_effectiveness}, and find that this modification does not lead to improved convergence or better downstream evaluation performance.

%% file: sections/method.tex
\subsection{Group reward-Decoupled normalization Policy Optimization}

To overcome these challenges, we propose \textbf{Group reward-Decoupled normalization Policy Optimization (GDPO)}, a method designed to better maintain distinctions among different reward combinations and more accurately capture their relative differences in the final advantages. 
In contrast to GRPO, which applies group-wise normalization directly to the aggregated reward sum, GDPO decouples this process by performing group-wise normalization of each reward separately before aggregation. Concretely, rather than summing all $n$ rewards first (as in Eq.~\ref{eq:grpo_sum}) and then applying group-wise normalization to obtain $A_{\text{sum}}$ (Eq.~\ref{eq:grpo_advantage}), GDPO computes the normalized advantage for each reward for the $j^{\text{th}}$ rollout of the $i^{\text{th}}$ question as: 
 \begin{equation}
 A^{(i,j)}_1 =
 \frac{r_1^{(i,j)} - \mathrm{mean}\{ r_1^{(i,1)}, \ldots, r_1^{(i,G)}\}}
 {\mathrm{std}\{ r_1^{(i,1)}, \ldots, r_1^{(i,G)}\}},
 \quad \ldots, \quad
 A^{(i,j)}_n =
 \frac{r_n^{(i,j)} - \mathrm{mean}\{ r_n^{(i,1)}, \ldots, r_n^{(i,G)}\}}
 {\mathrm{std}\{ r_n^{(i,1)}, \ldots, r_n^{(i,G)}\}}
 \end{equation}
 The overall advantage used for policy updates is then obtained by first summing the normalized advantages across all objectives:
\begin{gather}
 A_{\text{sum}}^{(i,j)} = A_1^{(i,j)} + \cdots + A_n^{(i,j)} \\
 \hat{A}^{(i,j)}_{\text{sum}} = 
\frac{
A^{(i,j)}_{\text{sum}}
- \mathrm{mean}\!\left\{
A^{(i',j')}_{\text{sum}}
\mid i' \in D_{\text{Batch}},\; j' = 1,\ldots,G
\right\}
}{
\mathrm{std}\!\left\{
A^{(i',j')}_{\text{sum}}
\mid i' \in D_{\text{Batch}},\; j' = 1,\ldots,G
\right\}
+ \epsilon
}
 \end{gather}
then applying batch-wise advantages normalization to the sum of the multi-reward advantages, which ensures that the numerical scale of the final advantage $\hat{A}^{(i,j)}_{\text{sum}}$ remains stable and does not grow as additional rewards are introduced. Empirically, we also find that this normalization step improves training stability, as shown in Appendix~\ref{sec:gdpo_wo_bn_fail_and_success}, where removing batch-wise normalization occasionally leads to convergence failures. 

By separating the normalization of each reward, GDPO alleviates the information-loss problem present in GRPO’s advantage estimation, as illustrated in Fig.~\ref{fig:advantage_illustration_grpo_gdpo}. Note that since the batch-wise normalization step in GDPO does not alter the number of distinct advantage groups, we omit it here for clarity. From the figure, we can see that when adopting GRPO, distinct reward combinations, such as $(0,2)$ and $(0,1)$, lead to identical normalized advantages, masking the subtle distinctions between them. In contrast, GDPO retains these fine-grained differences by assigning distinct advantage values to each combination, for example, the reward combination of $\mathtt{(0,1)}$ after GDPO normalization becomes $\mathit{(-0.7071,0.7071)}$ and $\mathtt{(0,2)}$ becomes $\mathit{(-1.4142,1.4142)}$, which more appropriately reflects that $\mathtt{(0,2)}$ should yield a stronger learning signal than $\mathtt{(0,1)}$. Similarly, when extending the number of rollouts to three, GRPO would assign advantage values of $\mathit{(0,0,0)}$ to $\mathtt{(1,1,1)}$. However, $\mathtt{(1,1,1)}$ may arise from heterogeneous reward partitions such as $r_1 = \mathtt{(1,1,0)}$ or $r_2 = \mathtt{(0,0,1)}$, for which GDPO would yield non-zero advantages, thereby preserving meaningful differences across reward dimension.

We further quantify the effectiveness of GDPO by comparing the number of distinct advantage groups across GDPO, GRPO, and GRPO w/o std under two experimental settings as shown in Fig.~\ref{fig:advantage_count_plot}.
In the two-reward scenario with a varying number of rollouts, GDPO consistently produces a significantly higher count of distinct advantage groups, with the gap widening as the number of rollouts increases. On the other hand, when fixing the number of rollouts to four and increasing the number of rewards, a similar pattern emerges, where GDPO exhibits progressively larger advantage granularity as the objective count grows. This demonstrate that the proposed decoupled normalization approach effectively increases the number of distinct advantage groups across all the RL settings and enables more precise advantage estimation. In addition to these theoretical improvements, we observe that using GDPO consistently yields a more stable training curve and improved convergence. For instance, GDPO achieves better convergence on both the format reward and the correctness reward in the tool-calling task, as shown in Fig.~\ref{fig:tool_training_curve}. GDPO also eliminates the training collapse issue observed with GRPO in the math reasoning task, as shown in Fig.~\ref{fig:math_1.5b_training_curve}, where the model trained with GDPO continues to improve the correctness reward score throughout training. Additional empirical results in Section~\ref{sec:exp} further confirm GDPO’s ability to achieve stronger alignment with the target preferences across a wide range of downstream tasks.

\subsection{Effective incorporation of priority variation}
\label{sec:how_to_incorporate}
Up to this point, we have assumed that all objectives carry equal importance. In practice, this assumption does not always hold in real-world applications. In this section, we provide a systematic overview of how to adjust the weights of rewards associated with different objectives or modify the reward functions to enforce prioritization of more important objectives. We also discuss how these two design choices behave differently when the underlying rewards vary significantly in difficulty. 

It is common practice to assign different weights to each reward to encode different priorities among objectives, such that $r_{\text{sum}} = w_1 r_1 + \dots + w_n r_n$, thereby controlling the contribution of each reward to the final advantage used for policy updates, and for GDPO, such weightings are apply to the normalized advantages of each reward as: 
\begin{equation}
 A_{\text{sum}}^{(i,j)} = w_1 A_1^{(i,j)} + \cdots + w_n A_n^{(i,j)}
 \end{equation}
However, we found that adjusting reward weights does not always yield the intended behavior when the difficulty levels of the underlying objectives differ substantially. If one objective is much easier than the others, the model often focuses on maximizing the reward for that objective regardless of the assigned weights. As a result, to more effectively force the model to allocate more attention to rewards associated with more challenging objectives, the weight differences must be made sufficiently large to compensate for the disparity in difficulty. Nevertheless, even with such adjustments, the model may still prefer optimizing the easier reward rather than the objective the user intends to prioritize, a phenomenon that we empirically demonstrate in Sec.~\ref{sec:different_reward_weight}.

Therefore, some recent works~\cite{liu2025laser,liu2025dler} address such reward hacking by conditioning easier rewards on more difficult rewards. Specifically, Given two rewards $r_k$ and $r_l$, conditioning $r_k$ on $r_l$ can be formulated as:
\begin{equation}
r_k =
\begin{cases}
r_k, & \text{if } r_l \geq t \\
0, & \text{otherwise.}
\end{cases}
\end{equation}
With such reward function design, the model can only receive reward of $r_k$ when the reward $r_l$ satisfies a predefined score threshold $t$, and as a consequence, the model get force to always maximize the human-prioritized reward first and completely alleviate the above mentioned problem. The empirical effectiveness of this strategy is shown in Sec.~\ref{sec:different_reward_weight}, where models trained with conditioned reward functions achieve higher performance on prioritized objectives compared to those trained without conditioning but only with larger weight assigned to the prioritized rewards. We also observe that, after resolving the issue of the easier reward dominating, assigning different reward weights for fine-grained priority adjustment can also be more faithfully reflected in the final model behavior. 

% In summary, adjusting reward weights is generally effective for capturing differences in human preferences. However, when one objective is considerably easier than the others, the model tends to maximize the easier reward regardless of its weight, reducing the effectiveness of modifying reward weights. In such cases, practitioners should instead modify the reward function by conditioning the easier objective on the prioritized one to better align the model with the desired behavior. We provide additional analysis in Sec.~\ref{sec:different_reward_weight} to illustrate the impact of these two techniques, namely adjusting reward weights and modifying the reward function.

%% file: sections/exp.tex
\subsection{Tool calling}
\label{sec:tool_calling}

\begin{figure}[h]
\begin{center}
\includegraphics[width=\textwidth]{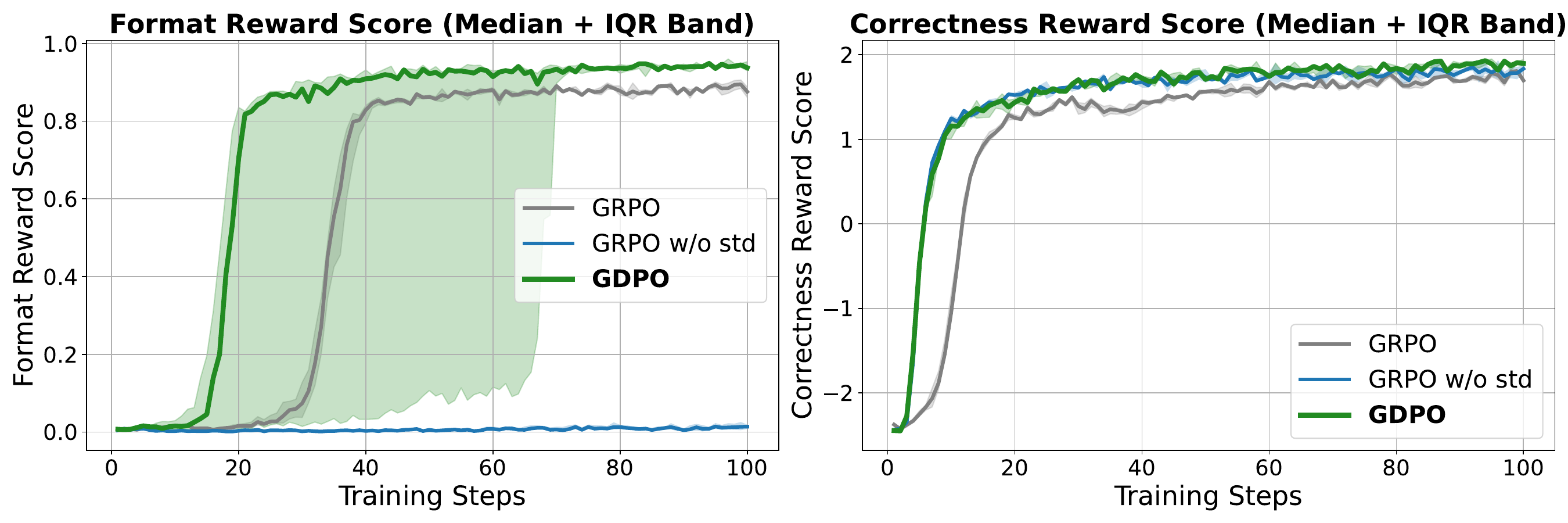}
\end{center}
\caption{Median and IQR reward curves across five runs of Qwen2.5-1.5B on the tool-calling task for GDPO, GRPO, and GRPO w/o std. GDPO consistently converges to higher correctness and format rewards, while GRPO w/o std matches correctness gains but fails to converge on the format reward.}
\label{fig:tool_training_curve}
\end{figure}

We compare GDPO with GRPO on the tool calling task following the setup of ToolRL~\cite{qian2025toolrl}\footnote{https://github.com/qiancheng0/ToolRL}. Specifically, the model is trained to learn how to incorporate external tools into the reasoning trajectory to solve a user task following the output format shown in Appendix~\ref{sec:toolrl_train_format}, where the reasoning steps must be enclosed in \texttt{<think></think>}, the tool calls must appear within \texttt{<tool\_call></tool\_call>}, and the model’s final answer must be placed inside \texttt{<response></response>}. We adopt the same training set as ToolRL for RL training, which consists of 2k samples from ToolACE~\cite{liu2024toolace}, 1k samples from Hammar~\cite{lin2024hammer}, and 1k samples from xLAM~\cite{zhang2025xlam}. Each training instance contains a question and its corresponding ground-truth tool calls. The training involves two rewards:
\begin{itemize}
    \item \textbf{Format reward}: The format reward $\mathcal{R}_{\text{format}} \in \{0,1\}$ checks whether the model output satisfies the required structure and contains all necessary fields in the correct order.
    \item \textbf{Correctness reward}: The correctness reward $\mathcal{R}_{\text{correct}} \in [-3,\,3]$ evaluates the model-generated tool calls against the ground-truth calls using three metrics: tool name matching, parameter name matching, and parameter content matching.
\end{itemize}
A full description of the reward formulation is provided in Appendix~\ref{sec:tool_calling_reward_function}. We train Qwen-2.5-Instruct (1.5B and 3B)~\cite{qwen2025qwen25technicalreport} with GRPO and GDPO using verl~\cite{sheng2024verl} for 100 steps, following the original hyperparameter settings from ToolRL’s GRPO recipe. We use four rollouts per training question, a batch size of 512, and a maximum response length of 1024. The complete hyperparameter configuration is listed in Appendix~\ref{sec:toolrl_hyperparameters}.

We evaluate the trained models on the Berkeley Function Call Leaderboard (BFCL-v3)~\cite{patilberkeley}, a comprehensive benchmark covering a broad range of challenges, including single-step reasoning, multi-step tool use, real-time execution, irrelevant tool rejection, simultaneous multi-tool selection, and multi-tool execution. We finetune the models with GRPO and GDPO across five runs and report the average accuracy and average format correctness on BFCL-v3 in Table~\ref{tab:tool_rl_main}. We additionally plot the median training curves with interquartile ranges for both methods over the five runs in Fig.~\ref{fig:tool_training_curve}. 

From the training curves, we observe that GDPO consistently converges to higher values on both the format and correctness reward score across all runs. Although GDPO exhibits larger variance in the number of steps required to converge on format reward, it ultimately attains better format compliance than both GRPO. For the correctness reward, GDPO shows faster early-stage improvement and reaches a higher reward score than the GRPO baselines toward later stages, demonstrating the effectiveness of GDPO on providing more accurate advantage estimation that leads to better optimization.

\begin{table}[!htp]\centering
\caption{Comparison of GDPO and GRPO-trained Qwen2.5-Instruct-1.5B/3B models on tool-calling accuracy and format correctness. The reported results are the averages across five runs.}\label{tab:tool_rl_main}
\resizebox{\textwidth}{!}{ % use this if the table is too large
\begin{tabular}{lrrr|rrr}\toprule
&Live Ocerall Acc ↑ &Multi Turn Overall Acc ↑ &Non-Live Overall Acc ↑ &\textbf{Avg Acc ↑} &\textbf{Correct Format ↑} \\\cmidrule{2-6}
Qwen2.5-Instruct-1.5B &37.89\% &0.12\% &15.63\% &17.88\% &4.74\% \\\cmidrule{1-6}
GRPO &50.63\% &2.04\% &37.87\% &30.18\% &76.33\% \\
GDPO &\textbf{55.36\%} & \textbf{2.50\%} &\textbf{40.58\%} &\textbf{32.81\%} &\textbf{80.66\%} \\\cmidrule{1-6}
Qwen2.5-Instruct-3B &63.57\% &1.38\% &30.75\% &31.90\% &58.37\% \\\cmidrule{1-6}
GRPO &69.23\% &3.14\% &45.24\% &39.20\% &81.64\% \\
GDPO &\textbf{71.22\%} &\textbf{4.59\%} &\textbf{46.79\%} &\textbf{40.87\%} &\textbf{82.23\%} \\\midrule
\bottomrule
\end{tabular}%
}
\end{table}

In the BFCL-v3 evaluation shown in Table~\ref{tab:tool_rl_main}, GDPO also consistently improves the average tool calling accuracy and format correctness over the GRPO-trained counterparts. For training Qwen2.5-Instruct-1.5B, GDPO achieves almost 5\% and 3\% improvement on Live/non-Live tasks and gains roughly 2.7\% improvement on the overall average accuracy and more than 4\% in correct format ratio compared with GRPO. Similar improvements are observed for the 3B model, where GDPO continues to outperform GRPO across all the sub-tasks, achieving up to 2\% accuracy improvement and delivers a better format compliance ratio.

\subsubsection{Does removing the standard deviation normalization term in GRPO provide any benefit?}
\label{sec:different_component_effectiveness}
\begin{table}[!htp]\centering
\caption{Comparison of GRPO, GRPO w/o std, and GDPO-trained Qwen2.5-Instruct-1.5B/3B models on tool-calling accuracy and format correctness.The reported results are the average across five runs. }\label{tab:grpo_wo_std_tool}
\resizebox{\textwidth}{!}{ % use this if the table is too large
\begin{tabular}{lrrr|rrr}\toprule
&Live Ocerall Acc ↑ &Multi Turn Overall Acc ↑ &Non-Live Overall Acc ↑ & \textbf{Avg Acc ↑} & \textbf{Correct Format ↑} \\\cmidrule{2-6}
Qwen2.5-1.5B-Instruct &37.89\% &0.12\% &15.63\% &17.88\% &4.74\% \\\cmidrule{1-6}
GRPO &50.63\% &2.04\% &37.87\% &30.18\% &76.33\% \\
GRPO w/o std &47.19\% &1.47\% &39.11\% &29.26\% & \textcolor{red}{0\%} \\
GDPO & \textbf{55.36\%} & \textbf{2.50\%} & \textbf{40.58\%} & \textbf{32.81\%} & \textbf{80.66\%} \\\midrule
\bottomrule
\end{tabular}%
}
\end{table}

Recall from Fig.~\ref{fig:advantage_count_plot} that removing the standard deviation normalization term in GRPO (denoted GRPO w/o std) slightly increases the number of distinct advantage groups. In this section, we empirically examine the effectiveness of this modification. Following the previous experiments, we run GRPO w/o std five times and report the average accuracy and average format correctness ratio on BFCL-v3. 

In the reward training curves shown in Fig.~\ref{fig:teaser_training_curve}, we observe that although GRPO w/o std converges to a correctness reward that is similar to GDPO and higher than standard GRPO, it fails to improve the format reward entirely. This failure results in a correct format ratio of 0\% on BFCL-v3 (see Table.~\ref{tab:grpo_wo_std_tool}), indicating that the model does not learn the required output structure. These also show that simply removing the standard deviation normalization term in order to increase advantage diversity can introduce instability into training, which may ultimately prevent successful convergence in multi-reward reinforcement learning.

% \finding{1}{The clipped tokens are often low-probability, high-entropy tokens that play a crucial role in exploration of reasoning paths and length control. Adopting a higher clipping threshold helps retain these tokens in gradient updates, thereby mitigating entropy collapse.}

\subsection{Mathematical reasoning}
\label{sec:math_reasoning}

We consider a mathematical reasoning task that optimizes two implicitly competing rewards: accuracy and adherence to a length constraint.
The goal is to improve model performance on challenging mathematical problems while keeping the generated output within a predefined response length to encourage efficient problem solving. We train DeepSeek-R1-1.5B, DeepSeek-R1-7B~\cite{guo2025deepseek}, and Qwen3-4B-Instruct~\cite{yang2025qwen3} using GRPO and GDPO on the DeepScaleR-Preview dataset~\cite{deepscaler2025} for 500 steps, which contains 40k competition-level math problems. Training is performed using verl~\cite{sheng2024verl}, and we follow the original DeepSeek-R1 prompt format~\cite{guo2025deepseek}. Following the DLER setup~\cite{liu2025dler}, we incorporate dynamic sampling, higher clipping thresholds, and the token-mean loss from DAPO~\cite{yu2025dapo}, and use 16 rollouts, a batch size of 512, and a maximum response length of 8000 tokens. The full set of hyperparameters is provided in Appendix~\ref{sec:math_code_reasoning_hyperparameters}.

The training uses two rewards:
\begin{itemize}
    \item \textbf{Length reward:} The length reward $\mathcal{R}_{\text{length}} \in \{0,1\}$ checks whether the model's output remains within the target length $l$, which is set to 4000 tokens for all remaining experiments:
    \[
        \mathcal{R}_{\text{length}} =
        \begin{cases}
        1, & \text{if response length} \leq l \\
        0, & \text{otherwise}.
        \end{cases}
    \]
    
    \item \textbf{Correctness reward:} The correctness reward $\mathcal{R}_{\text{correct}} \in \{0,1\}$ indicates whether the final answer extracted from the model's response matches the ground truth.
\end{itemize}

We compare the GRPO and GDPO-trained model on AIME-24~\cite{aime24}, AMC (AMC 2022 and AMC 2023)~\cite{amc}, MATH~\cite{hendrycks2021math500}, Minerva~\cite{lewkowycz2022minerva} and Olympiad Bench~\cite{he2024olympiadbench}. All evaluations are conducted using vLLM as the inference backend with a sampling temperature of 0.6, $top_p$ = 0.95, and a maximum response length of 32k tokens. For each evaluation question, we generate 16 samples and report the average pass@1 score and the average length-exceeding ratio, denoted Exceed, which measures the percentage of model responses that exceed the predefined length limit of 4000 tokens. 
\begin{figure}[h]
\begin{center}
\includegraphics[width=\textwidth]{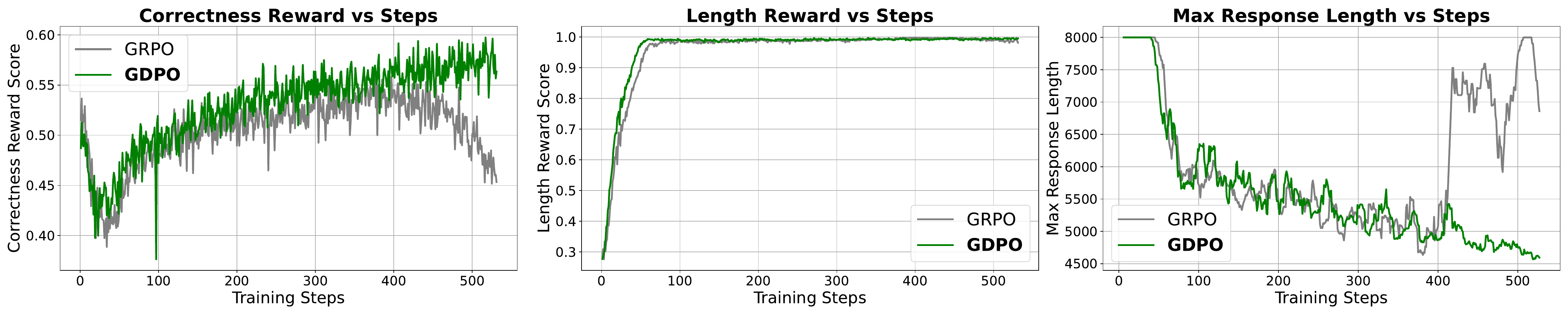}
\end{center}
\caption{Training behavior of GRPO and GDPO on DeepSeek-R1-1.5B across correctness reward, length reward, and maximum batch response length. Both methods rapidly maximize the length reward, briefly suppressing correctness, yet GDPO subsequently recovers it and surpasses GRPO. After roughly 400 steps, GRPO’s correctness score declines and its length-constraint violations increase, as reflected by rising maximum response lengths. In contrast, GDPO continues to improve correctness while steadily improving the control over response length.}
\label{fig:math_1.5b_training_curve}
\end{figure}

From the training curves of GRPO and GDPO on DeepSeek-R1-1.5B as shown in Fig.~\ref{fig:math_1.5b_training_curve}, we first observe that the model tends to maximize the easier reward regardless of the optimization method. In this case, the length reward is easier to optimize, and both GRPO and GDPO reach a full length score within roughly the first 100 training steps. We also see that this rapid rise in the length reward coincides with an early drop in the correctness reward, which indicates that the two rewards are competing. During the initial phase of training, the model prioritizes satisfying the length constraint, often at the expense of the more challenging correctness objective. However, from the correctness reward trajectories, we observe that GDPO recovers the correctness reward more effectively than GRPO, achieving higher correctness scores at comparable training steps. We also see that GRPO training starts to destabilize after 400 steps with the correctness rewards score gradually decreasing while GDPO continue to improve the correctness score. Moreover, although both GDPO and GRPO maintain nearly perfect length scores throughout training, we also record the maximum response length within each training batch to assess how well the models satisfy the length constraint under more extreme cases. The results show that, despite achieving almost a full length reward, the maximum response length for GRPO begins to increase sharply after roughly 400 training steps, while the maximum response length for GDPO continues to decrease. Similar observation can be seen on the training curves on DeepSeek-R1-7B and Qwen3-4B-Instruct as shown in Fig~\ref{fig:math_7b_training_curve} and Fig~\ref{fig:math_4b_training_curve} in appendix where we can see that GDPO consistently provide better alignment to the length constraint. This contrast further illustrates the effectiveness of GDPO in multi-reward optimization compared with GRPO.

% \begin{table}[!htp]\centering
% \caption{Comparison of GRPO/GDPO trained DeepSeek-R1-1.5B/7B models on Pass@1 accuracy and the proportion of responses exceeding the length constraint across mathematical reasoning benchmarks.}\label{tab:math_main_result}
% \resizebox{\textwidth}{!}{ % use this if the table is too large
% \begin{tabular}{lrrrrrrrrrr|rrr}\toprule
% &MATH ↑ &Exceed ↓ &AIME ↑ &Exceed ↓ &Amc ↑ &Exceed ↓ &Minerva ↑ &Exceed ↓ &Olympiad ↑ &Exceed ↓ & \textbf{Avg Acc ↑} & \textbf{Avg Exceed ↓}\\\cmidrule{2-13}
% DeepSeek-R1-1.5B &84.31 &34.98\% &29.79 &91.46\% &61.97 &67.47\% &38.41 &51.43\% &44.07 &70.10\% &51.71 &63.09\% \\\cmidrule{1-13}
% GRPO &83.57 &1.45\% &23.12 &10.83\% &64.5 &3.16\% &43.45 &1.68\% &44.31 &2.63\% &51.790 &3.95\% \\
% GDPO &\textbf{86.2} &\textbf{0.75\%} &\textbf{29.37} &\textbf{6.46\%} &\textbf{69.05} &\textbf{2.26\%} &\textbf{44.02} &\textbf{0.34\%} &\textbf{46.63} &\textbf{1.86\%} &\textbf{55.054} &\textbf{2.33\%} \\\cmidrule{1-13}
% DeepSeek-R1-7B &93.60 &26.02\% &55.40 &85.62\% &82.90 &57.23\% &49.79 &41.80\% &58.21 &60.57\% &67.98 &54.25\% \\\cmidrule{1-13}
% GRPO &\textbf{94.13} &0.47\% &50.2 &2.08\% &83.81 &0.60\% &53.2 &0.18\% &\textbf{60.19} &1.06\% &68.31 &0.88\% \\
% GDPO &93.9 &\textbf{0.10\%} &\textbf{53.12} &\textbf{0.21\%} &\textbf{83.96} &\textbf{0.30\%} &\textbf{53.81} &\textbf{0.07\%} &59.34 &\textbf{0.37\%} &\textbf{68.83} &\textbf{0.21\%} \\\midrule
% \bottomrule
% \end{tabular}%
% }
% \end{table}

\begin{table}[!htp]\centering
\caption{Comparison of GRPO and GDPO-trained DeepSeek-R1-1.5B/7B models on Pass@1 accuracy and the proportion of responses exceeding the length constraint across mathematical reasoning benchmarks.}\label{tab:math_main_result}
\resizebox{0.9\textwidth}{!}{ % use this if the table is too large
\begin{tabular}{lr|rrr|rrr|rrrr}\toprule
 &  &\multicolumn{3}{c}{DeepSeek-R1-1.5B} &\multicolumn{3}{c}{DeepSeek-R1-7B} &\multicolumn{3}{c}{Qwen3-4B-Instruct} \\\cmidrule{3-11}
 &  &- &GRPO &GDPO &- &GRPO &GDPO &- &GRPO &GDPO \\\cmidrule{1-2}\cmidrule{3-11}
\multirow{2}{*}[-0.3em]{MATH} & Acc ↑ &84.3\% &83.6\% &\textbf{86.2\%} &93.6\% &\textbf{94.1\%} &93.9\% &94.6\% & 93.9\% & 93.9\% \\\cmidrule{2-11}
&Exceed ↓ &35.0\% &1.5\% &\textbf{0.8\%} &26.0\% &0.5\% &\textbf{0.1\%} &11.3\% & 0.8\% & \textbf{0.1\%} \\\cmidrule{1-11}
\multirow{2}{*}[-0.3em]{AIME} & Acc ↑ &29.8\% &23.1\% &\textbf{29.4\%} &55.4\% &50.2\% &\textbf{53.1\%} &63.7\% &54.6\% &\textbf{56.9\%} \\\cmidrule{2-11}
&Exceed ↓ &91.5\% &10.8\% &\textbf{6.5\%} &85.6\% &2.1\% &\textbf{0.2\%} &71.3\% &\textbf{2.5\%} &\textbf{0.1\%} \\\cmidrule{1-11}
\multirow{2}{*}[-0.3em]{AMC} & Acc ↑ &62.0\% &64.5\% &\textbf{69.0\%} &82.9\% &83.8\% &\textbf{84.0\%} &84.5\% & \textbf{85.2\%} & 84.3\% \\\cmidrule{2-11}
&Exceed ↓ &67.5\% &3.2\% &\textbf{2.3\%} &57.2\% &0.6\% &\textbf{0.3\%} &33.9\% & 0.7\% & \textbf{0.1\%} \\\cmidrule{1-11}
\multirow{2}{*}[-0.3em]{Minerva} & Acc ↑ &38.41.\% &43.5\% &\textbf{44.0\%} &49.8\% &53.2\% &\textbf{53.8\%} &50.7\% & \textbf{52.4\%} & 51.9\% \\\cmidrule{2-11}
&Exceed ↓ &51.4\% &1.7\% &\textbf{0.3\%} &41.8\% &0.2\% &\textbf{0.1\%} &9.1\% & 0.3\% & \textbf{0.1\%} \\\cmidrule{1-11}
\multirow{2}{*}[-0.3em]{Olympiad} & Acc ↑ &44.1\% &44.3\% &\textbf{46.6\%} &58.2\% &\textbf{60.2\%} &59.7\% &65.7\% & 66.8\% & \textbf{67.5\%} \\\cmidrule{2-11}
&Exceed ↓ &70.1\% &2.6\% &\textbf{1.9\%} &60.6\% &1.1\% &\textbf{0.4\%} &41.3\% & 1.6\% & \textbf{1.0\%} \\\midrule
\bottomrule
\end{tabular}%
}
\end{table}

In addition, the benchmark results in Table~\ref{tab:math_main_result} show that the GDPO-trained models not only achieve substantial improvements in reasoning efficiency over the original models, with up to a 80\% reduction in length-exceeding ratios on AIME, but also deliver higher accuracy on the majority of the tasks. Moreover, GDPO generally outperforms GRPO on both the accuracy and length constraint objectives. For the DeepSeek-R1-1.5B, GDPO outperforms GRPO across all benchmarks, achieving accuracy improvements of 2.6\%/6.7\%/2.3\% on MATH, AIME and Olympiad, respectively, while also reducing the length exceed ratios across all the tasks. A similar trend holds for DeepSeek-R1-7B and Qwen3-4B-Instruct, where GDPO achieves stronger accuracy–efficiency trade-offs. The gains are particularly notable on the more challenging AIME benchmark, with GDPO improving accuracy by nearly 3\% while reducing the length-exceeding rate to 0.2\% and 0.1\%, compared with 2.1\% and 2.5\% under GRPO for DeepSeek-R1-7B and Qwen3-4B-Instruct. Together, these results show that GDPO not only improves reasoning accuracy across a range of mathematical tasks but also adheres to the length constraint more effectively, underscoring its advantage in multi-reward optimization.

\subsubsection{Impact analysis of different reward priority variation configurations}
\label{sec:different_reward_weight}
\begin{figure}[h]
\begin{center}
\includegraphics[width=\textwidth]{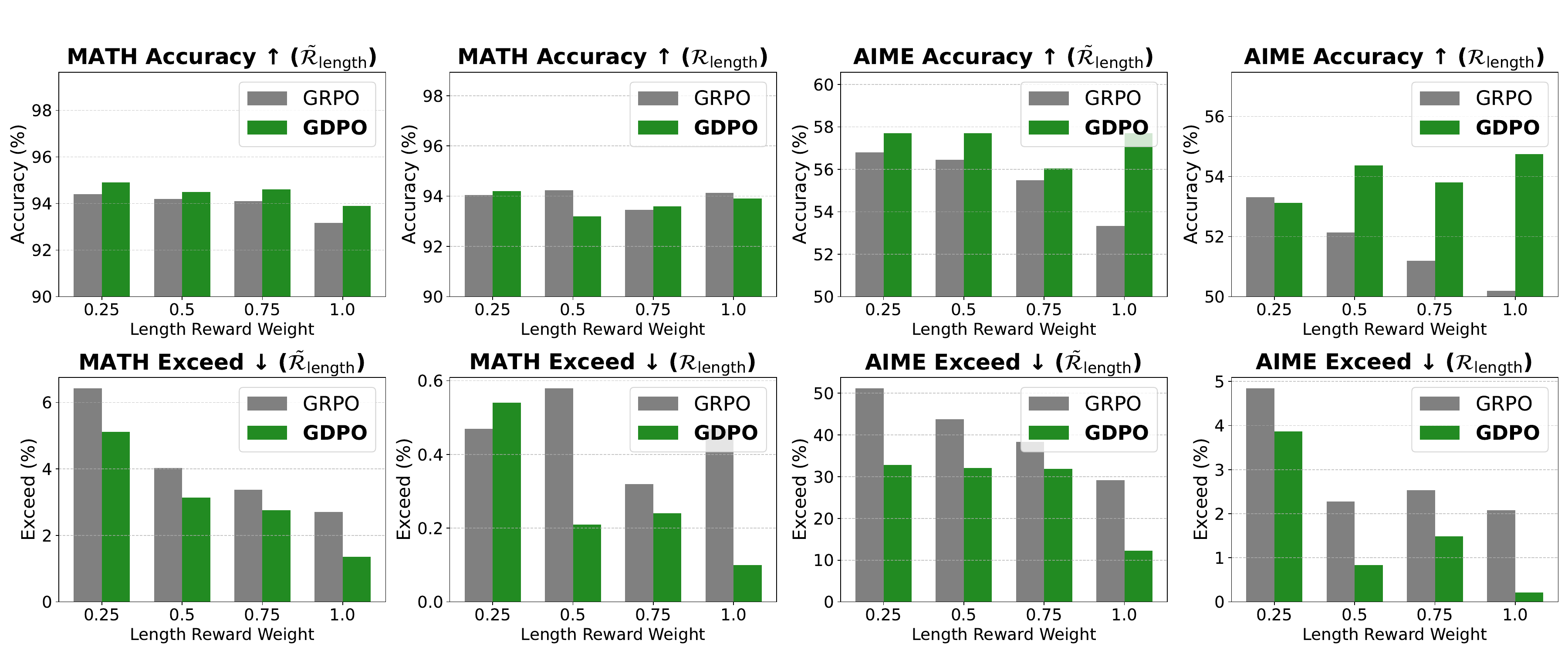}
\end{center}
\caption{Average accuracy and exceed-length ratios for GRPO/GDPO-trained DeepSeek-R1-7B models under varying length reward weights $\{1.0, 0.75, 0.5, 0.25\}$, with and without the conditioned length reward $\tilde{\mathcal{R}}_{\text{length}}$, on mathematical reasoning tasks.}
\label{fig:diff_weight_diff_human_prior_comparison}
\end{figure}

Until this point, we have assumed that all rewards are treated with equal priority. However, as shown in Fig.~\ref{fig:math_1.5b_training_curve}, the model often maximizes the easier objective at the cost of the more challenging one, even when both objectives are assigned the same reward weight. In this section, we investigate whether adjusting reward weights can guide the model to prioritize maximizing the correctness reward over the length reward when such a preference is desired and when the two objectives differ noticeably in difficulty.

We begin by fixing the reward weight for $\mathcal{R}_{\text{correct}}$, denoted $w_{\text{correct}}$, to 1, and varying the reward weight for $\mathcal{R}_{\text{length}}$, denoted $w_{\text{length}}$, over the set $\{0.25, 0.5, 0.75, 1.0\}$. This setup allows us to study whether reducing $w_{\text{length}}$ encourages the model to prioritize maximizing the more challenging correctness reward first. We carry out this experiment on DeepSeek-R1-7B and plot the average accuracy and average length-exceeding ratio of MATH and AIME in Fig.~\ref{fig:diff_weight_diff_human_prior_comparison}. Full results for the remaining tasks are provided in Appendix~\ref{sec:diff_weight_appendix}.

From the results, we observe that reducing $w_{\text{length}}$ to 0.75 or 0.5 has little impact on the average length-exceeding ratio, which shifts by only 0.4\% and 0.2\% for GRPO on AIME and by 1.3\% and 0.6\% for GDPO. In addition, lowering $w_{\text{length}}$ does not necessarily relax the length constraint, as decreasing $w_{\text{length}}$ from 0.75 to 0.5 does not consistently increase the length-exceeding ratio on either AIME or MATH for GRPO or GDPO. This suggests that simply adjusting reward weights does not reliably induce the intended prioritization when the underlying objectives differ substantially in difficulty. Only when $w_{\text{length}}$ is reduced to 0.25, making it sufficiently small to compensate for the difficulty gap between the objectives, do we observe a clear increase in the length-exceeding ratio on AIME for both GRPO and GDPO and on MATH for GDPO. 

% Moreover, we observe that model still tends to optimize the length reward first regardless of the chosen value of $w_{\text{length}}$, highlighting that reward weighting alone does not fully resolve the difficulty imbalance between objectives. 

\begin{figure}[h]
\begin{center}
\includegraphics[width=\textwidth]{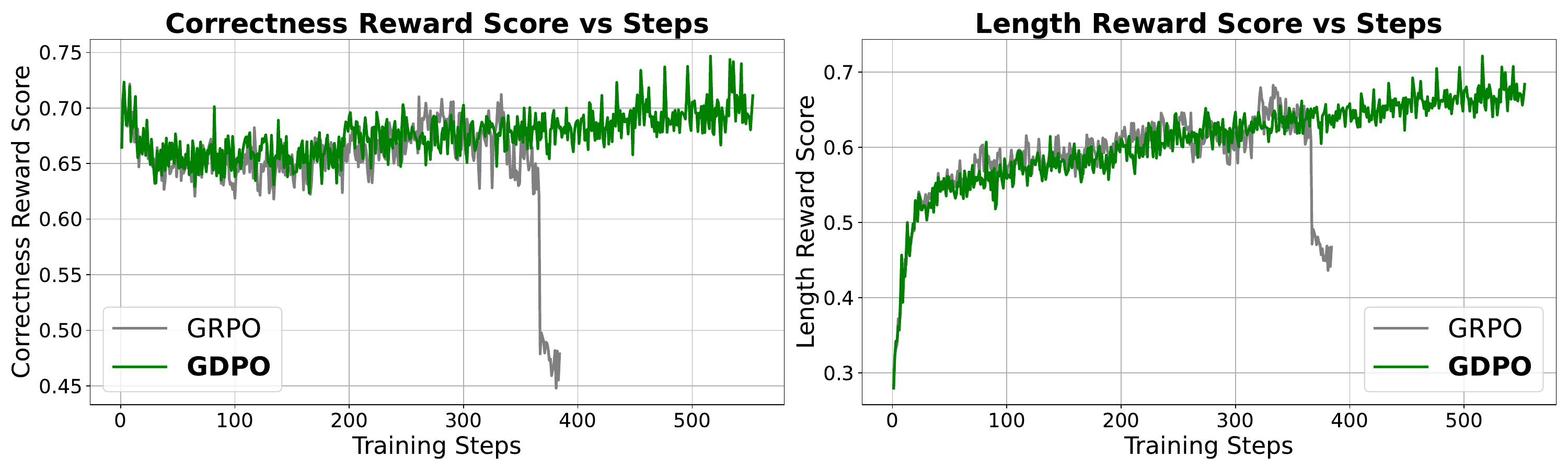}
\end{center}
\caption{Training curves of GRPO and GDPO with the conditioned length reward $\tilde{\mathcal{R}}_{\text{length}}$ on DeepSeek-R1-7B across correctness reward, length reward.}
\label{fig:math_7b_human_prior_training_curve}
\end{figure}

We next investigate whether conditioning the easier length reward on the more challenging correctness reward can help mitigate the disparity in difficulty between the two objectives and help improve priority alignment. Following the formulation in Sec.~\ref{sec:how_to_incorporate}, we replace the original length reward $\mathcal{R}_{\text{length}}$ with a conditioned length reward defined as:
    \[     \mathcal{\tilde{R}}_{\text{length}} =
        \begin{cases}
        1, & \text{if response length} \leq l \textcolor{red}{\text{ and } \mathcal{R}_{\text{correct}} = 1} \\
        0, & \text{otherwise}.
        \end{cases}
    \]
Under this formulation, the model receives the length reward only when the generated response is also correct.

\begin{table}[!htp]\centering
\caption{Comparison of GRPO and GDPO DeepSeek-R1-7B models, with and without the conditioned length reward $\tilde{\mathcal{R}}_{\text{length}}$, on Pass@1 accuracy and the ratio of outputs exceeding the length constraint across mathematical reasoning benchmarks.}\label{tab:adopt_new_lengt_reward}
\resizebox{0.6\textwidth}{!}{ % use this if the table is too large
\begin{tabular}{lr|rrr|rrr}\toprule
& &\multicolumn{5}{c}{DeepSeek-R1-7B} \\\cmidrule{3-7}
& &- &\multicolumn{2}{c|}{$\mathcal{R}_{\text{length}}$} &\multicolumn{2}{c}{$\tilde{\mathcal{R}}_{\text{length}}$} \\\cmidrule{3-7}
& &- &GRPO &GDPO &GRPO &GDPO \\\cmidrule{1-7}
\multirow{2}{*}[-0.3em]{MATH} & Acc ↑ &93.6\% &\textbf{94.1\%} &93.9\% &93.2\% &\textbf{93.9\%} \\\cmidrule{2-7}
&Exceed ↓ &26.0\% &0.5\% &\textbf{0.1\%} &2.7\% &\textbf{1.4\%} \\\cmidrule{1-7}
\multirow{2}{*}[-0.3em]{AIME} & Acc ↑ &55.4\% &50.2\% &\textbf{53.1\%} &53.3\% &\textbf{57.7\%} \\\cmidrule{2-7}
&Exceed ↓ &85.6\% &2.1\% &\textbf{0.2\%} &29.2\% &\textbf{12.3\%} \\\cmidrule{1-7}
\multirow{2}{*}[-0.3em]{AMC} & Acc ↑ &82.9\% &83.8\% &\textbf{84.0\%} &82.9\% &\textbf{85.9\%} \\\cmidrule{2-7}
&Exceed ↓ &57.2\% &0.6\% &\textbf{0.3\%} &8.6\% &\textbf{3.8\%} \\\cmidrule{1-7}
\multirow{2}{*}[-0.3em]{Minerva} & Acc ↑ &49.8\% &53.2\% &\textbf{53.8\%} &53.2\% &\textbf{53.4\%} \\\cmidrule{2-7}
&Exceed ↓ &41.8\% &0.2\% &\textbf{0.1\%} &2.5\% &\textbf{1.0\%} \\\cmidrule{1-7}
\multirow{2}{*}[-0.3em]{Olympiad} & Acc ↑ &58.2\% &\textbf{60.2\%} &59.7\% &59.1\% &\textbf{60.8\%} \\\cmidrule{2-7}
&Exceed ↓ &60.6\% &1.1\% &\textbf{0.4\%} &10.3\% &\textbf{6.2\%} \\\midrule
\bottomrule
\end{tabular}%
}
\end{table}

First, we observe that adopting the modified reward function $\tilde{\mathcal{R}}_{\text{length}}$ prevents the model from aggressively maximizing the length reward at the start of training. This reward design also helps avoid large drops in correctness reward score as the model tries to satisfy the length constraint. We can see that the average correctness reward decreases only slightly early in training and then gradually recovers from Fig.~\ref{fig:math_7b_human_prior_training_curve}.

From Table~\ref{tab:adopt_new_lengt_reward}, we also observe that using 
$\tilde{\mathcal{R}}_{\text{length}}$ leads to a larger increase in the average length-exceeding ratio for both GRPO and GDPO compared with merely adjusting the weight $w_{\text{length}}$ of $\mathcal{R}_{\text{length}}$, indicating a more effective relaxation of the length constraint. However, GRPO fails to convert this relaxed constraint into meaningful accuracy improvements. In contrast, GDPO prioritizes the correctness reward more effectively and achieves more consistent accuracy improvement over training without $\tilde{\mathcal{R}}_{\text{length}}$, while introducing substantially smaller increases in length violations. For instance, using $\tilde{\mathcal{R}}{\text{length}}$ with GDPO yields a 4.4\% accuracy improvement on AIME with a 16.9\% reduction in length-exceeding ratio, and a 3\% accuracy gain on AMC with a 4.8\% reduction in length violations compared with using GRPO with the same reward.

We next examine whether, after mitigating the difficulty disparity through conditioned length reward, varying the reward weight on $\tilde{\mathcal{R}}_{\text{length}}$, denoted $\tilde{w}_{\text{length}}$, leads to more faithful reflection of fine-grained preference adjustments. We fix the correctness reward weight and vary $\tilde{w}_{\text{length}} \in \{0.25, 0.5, 0.75, 1.0\}$. As shown in Fig.~\ref{fig:diff_weight_diff_human_prior_comparison}, the models trained with conditioned reward behave more predictably. For example, reducing $\tilde{w}_{\text{length}}$ from 1.0 to 0.25 steadily increases the length-exceeding ratio for both GRPO and GDPO on MATH and AIME, in contrast to the unstable results observed when adjusting the weight of the original $\mathcal{R}_{\text{length}}$.

Finally, across all settings, including different reward formulations and different reward weights, GDPO consistently provides a better accuracy and efficiency trade-off than GRPO.

\subsection{Coding reasoning}
\label{sec:coding_reasoning}

\begin{table}[!htp]\centering
\caption{Comparison of GRPO and GDPO trained DeepSeek-R1-7B models on coding pass rate, length-exceeding rate, and bug ratio across coding reasoning benchmarks. Here, $\mathcal{R}_{\text{pass}} + \tilde{\mathcal{R}}_{\text{length}}$ refers to optimizing $\mathcal{R}_{\text{pass}}$ and $\tilde{\mathcal{R}}_{\text{length}}$, and $\mathcal{R}_{\text{pass}} + \tilde{\mathcal{R}}_{\text{length}} +\mathcal{R}_{\text{bug}} $ refers to optimizing $\mathcal{R}_{\text{pass}}$, $\tilde{\mathcal{R}}_{\text{length}}$, and $\mathcal{R}_{\text{bug}}$.}\label{tab:coding_reaspning_main}
\resizebox{0.7\textwidth}{!}{ % use this if the table is too large
\begin{tabular}{lr|ccc|cll}\toprule
& &\multicolumn{5}{c}{DeepSeek-R1-7B} \\\cmidrule{3-7}
& &- &\multicolumn{2}{c|}{$\mathcal{R}_{\text{Pass}} + \mathcal{\tilde{R}}_{\text{length}}$} &\multicolumn{2}{c}{$\mathcal{R}_{\text{Pass}} + \mathcal{\tilde{R}}_{\text{length}} + \mathcal{R}_{\text{Bug}}$} \\\cmidrule{3-7}
& &- &$\text{GRPO}_{\text{2-obj}}$ &$\text{GDPO}_{\text{2-obj}}$ &$\text{ }$$\text{ }$$\text{ }$$\text{ }$$\text{GRPO}_{\text{3-obj}}$ &$\text{GDPO}_{\text{3-obj}}$ \\\cmidrule{1-7}
\multirow{4}{*}{Apps} &Pass ↑ &28.1\% &67.2\% &\textbf{68.3\%} &$\text{ }$$\text{ }$$\text{ }$$\text{ }$\textbf{68.1\%} &67.8\% \\\cmidrule{2-7}
&Exceed ↓ &73.9\% &5.2\% &\textbf{5.0\%} &$\text{ }$$\text{ }$$\text{ }$$\text{ }$11.2\% &\textbf{8.5\%} \\\cmidrule{2-7}
&Bug ↓ &32.9\% &25.0\% &\textbf{23.5\%} &$\text{ }$$\text{ }$$\text{ }$$\text{ }$20.3\% &\textbf{18.8\%} \\\cmidrule{1-7}
\multirow{4}{*}{Codecontests} &Pass ↑ &47.3\% &63.2\% &\textbf{65.8\%} &$\text{ }$$\text{ }$$\text{ }$$\text{ }$\textbf{65.6\%} &\textbf{65.6\%} \\\cmidrule{2-7}
&Exceed ↓ &83.0\% &\textbf{14.2\%} &14.3\% &$\text{ }$$\text{ }$$\text{ }$$\text{ }$19.3\% &\textbf{15.8\%} \\\cmidrule{2-7}
&Bug ↓ &29.7\% &14.1\% &\textbf{13.2\%} &$\text{ }$$\text{ }$$\text{ }$$\text{ }$3.9\% &\textbf{2.5\%} \\\cmidrule{1-7}
\multirow{4}{*}{Codeforces} &Pass ↑ &46.5\% &68.1\% &\textbf{71.2\%} &$\text{ }$$\text{ }$$\text{ }$$\text{ }$\textbf{69.5\%} &69.4\% \\\cmidrule{2-7}
&Exceed ↓ &82.8\% &\textbf{18.1\%} &18.4\% &$\text{ }$$\text{ }$$\text{ }$$\text{ }$16.9\% &\textbf{13.6\%} \\\cmidrule{2-7}
&Bug ↓ &27.8\% &7.0\% &\textbf{5.6\%} &$\text{ }$$\text{ }$$\text{ }$$\text{ }$2.5\% &\textbf{1.8\%} \\\cmidrule{1-7}
\multirow{4}{*}{Taco} &Pass ↑ &28.1\% &45.1\% &\textbf{48.4\%} &$\text{ }$$\text{ }$$\text{ }$$\text{ }$44.4\% &\textbf{45.1\%} \\\cmidrule{2-7}
&Exceed ↓ &78.0\% &11.8\% &\textbf{10.8\%} &$\text{ }$$\text{ }$$\text{ }$$\text{ }$14.7\% &\textbf{10.6\%} \\\cmidrule{2-7}
&Bug ↓ &48.9\% &37.7\% &\textbf{36.2\%} &$\text{ }$$\text{ }$$\text{ }$$\text{ }$30.0\% &\textbf{28.0\%} \\\midrule
\bottomrule
\end{tabular}%
}
\end{table}

We examine whether GDPO continues to outperform GRPO when optimizing more than two rewards on our coding reasoning task. Similar to the mathematical reasoning setup, the objective is to improve the model's coding performance while constraining its output to a predefined target length. In addition, we introduce a third objective that encourages the model to generate bug-free code. We compare GDPO and GRPO by training DeepSeek-R1-7B on the Eurus-2-RL dataset~\cite{cui2025process}, which contains 24k coding problems, each with multiple test cases. Training is conducted using the verl~\cite{sheng2024verl} framework for 400 steps, and we adopt the same hyperparameter configuration used in the mathematical reasoning experiments. The training optimizes three rewards:

\begin{itemize}
    \item \textbf{Passrate reward:} The passrate reward $\mathcal{R}_{\text{pass}} \in [0,1]$ measures the proportion of test cases passed by the generated code:
        \[
            \mathcal{R}_{\text{pass}} =
            \frac{\text{number of passed test cases}}{\text{total test cases}}.
        \]
    \item \textbf{Conditioned Length reward:} The length reward $\mathcal{\tilde{R}}_{\text{length}}\in \{0,1\}$ checks whether the model’s response remains within the target length $l$ and whether the generated code satisfies correctness requirements:
    \[
        \mathcal{\tilde{R}}_{\text{length}} =
        \begin{cases}
        1, & \text{if response length } \leq l \text{ and } \mathcal{R}_{\text{pass}} = 1, \\
        0, & \text{otherwise}.
        \end{cases}
    \]

    \item \textbf{Bug reward:} The bug reward $\mathcal{R}_{\text{bug}} \in \{0,1\}$ indicates whether the generated code runs without runtime or compilation errors.
\end{itemize}

For evaluation, we assess the trained model on the validation set from PRIME~\cite{cui2025process}, which includes Apps~\cite{hendrycks2021measuring}, CodeContests~\cite{li2022competition}, Codeforces\footnote{https://huggingface.co/datasets/MatrixStudio/Codeforces-Python-Submissions}, and Taco~\cite{li2023taco}. Following the same settings used for the mathematical reasoning evaluations, we use a sampling temperature of 0.6, a $top_p$ value of 0.95, and a maximum response length of 32k tokens. For each evaluation question, we generate 16 rollouts and report the average test case pass rate, the average length-exceeding ratio, and the average bug ratio, where the bug ratio measures the proportion of generated code that results in either a runtime error or a compilation error. 

We compare GDPO and GRPO under two configurations: (1) a two-reward setting using $\mathcal{R}_{\text{pass}}$ and $\tilde{\mathcal{R}}_{\text{length}}$, and (2) a three-reward setting using $\mathcal{R}_{\text{pass}}$, $\tilde{\mathcal{R}}_{\text{length}}$, and $\mathcal{R}_{\text{bug}}$. We denote the two-reward and three-reward versions of GRPO as $\text{GRPO}_{\text{2-obj}}$ and $\text{GRPO}_{\text{3-obj}}$, and use the same notation for GDPO. As shown in Table~\ref{tab:coding_reaspning_main}, $\text{GDPO}_{\text{2-obj}}$ improves pass rates across all the tasks compared with $\text{GRPO}_{\text{2-obj}}$, while maintaining a similar length-exceeding ratio. For example, $\text{GDPO}_{\text{2-obj}}$ improves the Codecontests pass rate by 2.6\% while increasing the length-exceeding ratio by only 0.1\%, and achieves a 3.3\% pass rate gain together with a 1\% reduction in length violations compared with $\text{GRPO}_{\text{2-obj}}$ on Taco. A similar pattern holds in the three-reward setting where $\text{GDPO}_{\text{3-obj}}$ achieves a substantially better balance across all objectives, maintaining similar pass rate to $\text{GRPO}_{\text{3-obj}}$ while also markedly reducing both the length-exceeding ratio and the bug ratio.

Overall, these results demonstrate that GDPO remains effective as the number of reward signals increases. It consistently achieves a more favorable trade-off across all objectives than GRPO in both the two-reward and three-reward configurations.